\documentclass{article} %
\usepackage{iclr2027_conference,times}

\usepackage{amsmath,amsfonts,bm}

\def\Tabref#1{Tab.~\ref{#1}}

\def\Figref#1{Fig.~\ref{#1}}

\def\Secref#1{Sec.~\ref{#1}}

\def\eqref#1{eq.~\ref{#1}}
\def\Eqref#1{Eq.~\ref{#1}}

\def\1{\bm{1}}

\DeclareMathAlphabet{\mathsfit}{\encodingdefault}{\sfdefault}{m}{sl}
\SetMathAlphabet{\mathsfit}{bold}{\encodingdefault}{\sfdefault}{bx}{n}

\newcommand{\KL}{D_{\mathrm{KL}}}

\usepackage[utf8]{inputenc} %
\usepackage[T1]{fontenc}    %
\usepackage{hyperref}       %
\usepackage{url}            %
\usepackage{amsfonts}       %
\usepackage{nicefrac}       %

\usepackage{amsmath}
\usepackage{amssymb}
\usepackage{amsthm}    %
\usepackage{dsfont}
\usepackage{booktabs}
\usepackage{placeins}
\usepackage{dsfont}
\usepackage{graphicx}
\usepackage{makecell}
\usepackage{multirow}
\usepackage{pifont}%
\usepackage[group-separator={,}]{siunitx}
\usepackage{tabularx}

\usepackage{xcolor}
\usepackage{caption}
\usepackage{soul}
\usepackage{changepage,threeparttable}
\usepackage{array}
\usepackage{colortbl}
\usepackage{algpseudocode}
\usepackage{algorithm}
\usepackage{mathtools}
\usepackage{xfrac}

\usepackage{catchfile}
\usepackage{longtable}
\usepackage{geometry}
\usepackage{enumitem}
\usepackage{tabu}
\usepackage{supertabular}
\usepackage{multicol}
\usepackage{algpseudocode}
\usepackage{algorithm}
\usepackage{wrapfig}
\usepackage{graphicx}
\usepackage{caption}
\usepackage{subcaption}
\usepackage{adjustbox}
\newcolumntype{Y}{>{\centering\arraybackslash}X}

\definecolor{color_1}{RGB}{255,0,128}
\definecolor{color_2}{RGB}{0,128,128}
\definecolor{color_3}{RGB}{0,128,0}
\definecolor{color_4}{RGB}{128,0,0}
\definecolor{color_5}{RGB}{128,0,128}
\definecolor{cadetgrey}{RGB}{0.57, 0.64, 0.69}
\definecolor{our_color}{RGB}{0,65,145}

\newtheorem{proposition}{Proposition}
\newtheorem{theorem}{Theorem}
\newtheorem{corollary}{Corollary}

\usepackage{amssymb}%
\usepackage{pifont}%

\newcommand{\methodname}{UW-Flow}
\usepackage{graphicx}
\usepackage{tikz}

\newcommand{\uotnote}[1]{%
    \hfill{\footnotesize\color{gray!80}$\triangleright$\,#1}}

\usepackage{graphicx}
\usepackage{booktabs}
\usepackage{caption}
\usepackage{cuted}

\newsavebox{\imagenetTableBox}
\newsavebox{\imagenetQualBox}


\definecolor{correctblue}{RGB}{0, 92, 230}
\definecolor{wrongred}{RGB}{243,1,3}

\title{Unbalanced W-Flow: Unbalanced Wasserstein Drifting under Finite-Batch Marginal Fluctuations}

\author{%
\parbox[t]{\dimexpr\textwidth-2\tabcolsep\relax}{\centering
\begin{tabular}{@{}*{3}{>{\centering\arraybackslash}p{0.3\linewidth}}@{}}
Jongwook Jeon$^{1}$ & Mingue Park$^{2}$ & Jiwon Choi$^{2}$ \\[8pt]
Jaihoon Kim$^{2}$ & Yunhong Min$^{2}$ & Minhyuk Sung$^{2}$
\end{tabular}\\[10pt]
{\normalfont $^{1}$Yonsei University \qquad $^{2}$KAIST}\\[4pt]
\mbox{\normalfont\fontsize{8pt}{10pt}\selectfont\texttt{jjwook2602@yonsei.ac.kr,\ \{kicikicik, jiwonch, jh27kim, dbsghd363, mhsung\}@kaist.ac.kr}}
}}

\iclrfinalcopy %
\begin{document}
\maketitle
\lhead{Preprint} %

\begin{abstract}
\vspace{-0.5\baselineskip}
Teacher-free one-step generative models such as W-Flow learn directly from distributional objectives, with the Sinkhorn divergence providing the strongest empirical performance among the energies studied in W-Flow. However, Sinkhorn divergence is based on balanced optimal transport, whose exact marginal constraints can induce undesirable transport under finite minibatch fluctuations. We introduce \textbf{\methodname}, which replaces the balanced Sinkhorn energy with a debiased unbalanced Sinkhorn divergence and derives the corresponding Wasserstein velocity. The resulting dynamics retain W-Flow's cross-minus-self structure, recover the balanced limit, and preserve target identifiability under suitable assumptions. We further characterize the finite-batch velocity error of balanced and unbalanced transport by comparing each minibatch field with its population counterpart. Our analysis identifies conditions under which marginal relaxation yields smaller minibatch-to-population velocity error, and shows that for separated matched modes, balanced sensitivity grows with inter-mode separation while the corresponding Unbalanced Optimal Transport (UOT) sensitivity remains controlled by local transport geometry. Controlled experiments on fixed-support and continuous two-mode distributions support these predictions. On MNIST, CIFAR-10, and ImageNet-100, \methodname{} improves generation under matched training budgets, with the largest gains in unconditional generation and consistent improvements in the class-conditional setting.
\vspace{-0.5\baselineskip}

\end{abstract}

\section{Introduction}
\label{sec:intro}
\vspace{-0.5\baselineskip}
Recent progress in teacher-free one-step generative modeling has shown that high-quality generators can be learned directly from distributional objectives~\citep{zhou:2025imm, deng:2026drifting, han:2026wflow}. W-Flow~\citep{han:2026wflow} is a representative example that defines a Wasserstein gradient flow from a distributional energy and trains a one-step generator to follow the resulting particle updates. Among the distributional energies studied in W-Flow, the Sinkhorn divergence~\citep{genevay:2018learning,feydy:2019sinkhorn} achieves the strongest empirical performance.

The Sinkhorn divergence used by W-Flow is built on \emph{balanced} optimal transport~\citep{cuturi:2013sinkhorn}, whose coupling must exactly reproduce both input marginals. In practice, however, the corresponding velocity field is estimated from finite minibatches~\citep{zhu:2024neural,han:2026wflow}, whose empirical composition fluctuates from one update to another. Exact marginal matching can therefore force transport between geometrically unrelated regions when, for example, the sampled proportions of different modes do not agree~\citep{fatras:2021uot,nguyen:2022improving}. Unbalanced optimal transport (UOT) relaxes these hard marginal constraints through soft penalties~\citep{liero:2018optimal,chizat:2018uot}, allowing the transported marginals to deviate when exact matching would require undesirable transport. Prior work has shown that such marginal relaxation can improve robustness to minibatch composition mismatch~\citep{fatras:2021uot,nguyen:2022improving} and outliers~\citep{balaji:2020robust}.

Motivated by this, we introduce \textbf{UW-Flow}, which uses the debiased unbalanced Sinkhorn divergence~\citep{sejourne:2019sinkhornuot} as its distributional energy, and derive the resulting Wasserstein velocity. In contrast to prior uses of UOT for minibatch objectives~\citep{fatras:2021uot}, transport maps or couplings~\citep{yang:2018scalable,choi:2023semidualuot,eyring:2024unbalancedness}, and scalable JKO schemes~\citep{choi:2024scalablewgf}, our formulation uses the debiased unbalanced Sinkhorn divergence as the population energy that directly defines W-Flow's one-step velocity. In W-Flow, the transport objective directly determines the velocity field used for one-step learning, and thus replacing the balanced Sinkhorn energy changes the underlying population dynamics rather than merely modifying a coupling within a fixed objective. We show that UW-Flow retains the cross-minus-self structure of W-Flow, recovers the balanced dynamics in the hard-marginal limit~\citep{sejourne:2019sinkhornuot}, and preserves the desired target-identification property under suitable assumptions. Marginal relaxation therefore defines alternative Wasserstein dynamics while retaining the key structural properties required for distribution matching.

We next characterize how finite-batch sampling perturbs the velocity fields induced by balanced and unbalanced transport. Specifically, we measure how far each minibatch velocity deviates from its own population counterpart. For balanced Sinkhorn transport, exact marginal preservation imposes an unavoidable finite-batch velocity error. For UW-Flow, we characterize the leading sensitivity to empirical marginal perturbations and identify conditions under which the UOT minibatch-to-population velocity error is smaller than the corresponding balanced error. In particular, for separated matched modes, the balanced sensitivity grows with inter-mode separation, whereas the corresponding UOT sensitivity remains controlled by the local transport geometry. We further test these predictions on fixed-support and continuous two-mode toy distributions by varying batch size, mode separation, and marginal relaxation.

We evaluate UW-Flow on MNIST~\citep{lecun:1998mnist}, CIFAR-10~\citep{krizhevsky:2009cifar10}, and ImageNet-100~\citep{deng:2009imagenet,tian:2020cmc} under matched training budgets. The gains are largest in unconditional generation, where multimodal composition mismatch is more substantial, while class-conditional generation also benefits from marginal relaxation.

\section{Related Work}
\label{sec:related}

\vspace{-0.5\baselineskip}
\paragraph{Wasserstein flows for one-step generation.}
Wasserstein gradient flows provide generative models with particle velocities induced by a distributional energy. NSGF~\citep{zhu:2024neural} studies flows driven by the Sinkhorn divergence, while W-Flow~\citep{han:2026wflow} trains a one-step generator to regress onto the resulting particle updates. Optimal transport (OT) has also been used as a training loss~\citep{genevay:2018learning}, to construct generators~\citep{rout:2021generativeotm}, and to pair samples for flow matching~\citep{Tong:2024batchot,pooladian:2023multisamplefm}. In W-Flow, the velocity is estimated from balanced minibatch couplings rather than population distributions. Because these couplings must match the sampled marginals exactly, fluctuations in regional sample proportions can alter the transport plan~\citep{fatras:2019learning, fatras:2021uot, nguyen:2021transportation, nguyen:2022improving} and perturb the velocity used as the regression target.

\vspace{-0.5\baselineskip}
\paragraph{Relaxed transport.}
Unbalanced transport relaxes exact marginal matching and has been used to improve the robustness of minibatch couplings~\citep{sejourne:2019sinkhornuot,fatras:2021uot}. Related approaches use partial transport to limit mismatched minibatch pairings~\citep{nguyen:2022improving}, learn maps with reweighted marginals~\citep{choi:2023semidualuot}, incorporate UOT into flow matching~\citep{eyring:2024unbalancedness}, or formulate scalable JKO-based flows~\citep{choi:2024scalablewgf}. We incorporate an unbalanced Sinkhorn divergence into the W-Flow~\citep{han:2026wflow} framework while retaining the target distribution as an equilibrium of the induced velocity field.

\vspace{-0.5\baselineskip}
\paragraph{Statistical analysis of transport estimates.}
Prior analyses study empirical entropic OT maps~\citep{rigollet:2025sample,goldfeld:2024limit} and minibatch OT objectives~\citep{fatras:2019learning,fatras:2021uot}. \citet{fatras:2021uot} also demonstrate minibatch UOT particle flows, but their concentration and gradient results concern the expected minibatch loss. To our knowledge, none directly compares the minibatch-to-population error of the cross-minus-self velocity under balanced and unbalanced marginal constraints. We make this comparison and identify a separated-mode setting in which the leading unbalanced error is smaller.

\section{Preliminaries}
\label{sec:preliminary}
\vspace{-0.5\baselineskip}

We first review Wasserstein gradient flows as a framework for defining distributional dynamics and particle updates (\Secref{subsec:wgf}). We then describe how W-Flow instantiates this framework with the balanced Sinkhorn divergence and learns the resulting updates with a one-step generator (\Secref{subsec:balanced_sinkhorn}).

\subsection{Wasserstein Dynamics Induced by Distributional Energies}
\label{subsec:wgf}
\vspace{-0.5\baselineskip}

Let $p\in\mathcal{P}_2(\mathbb{R}^d)$ be a target distribution and $q\in\mathcal{P}_2(\mathbb{R}^d)$ an evolving distribution, where $\mathcal{P}_2(\mathbb{R}^d)$ denotes the space of probability measures with finite second moment. Given a differentiable distributional energy $\mathcal{F}(q;p)$ measuring the discrepancy between $q$ and $p$, its formal $2$-Wasserstein gradient flow~\citep{ambrosio:2005gradientflows} satisfies
\begin{equation}
\partial_t q_t
=
\nabla\cdot\left(
q_t\nabla\frac{\delta\mathcal{F}}{\delta q}(q_t;p)
\right).
\label{eq:wgf}
\end{equation}
Equivalently, this evolution satisfies the continuity equation
\begin{equation}
\partial_t q_t+\nabla\cdot(q_tV_t)=0,
\qquad
V_t(x)
=
-\nabla_x\frac{\delta\mathcal{F}}{\delta q}(q_t;p)(x).
\label{eq:continuity}
\end{equation}
The choice of energy thus determines the velocity field through which the distribution evolves. Under suitable regularity conditions, the same dynamics can be described at the particle level. A particle initialized at $X_0=x$ follows
\begin{equation}
\frac{dX_t}{dt}
=
V_{q_t,p}(X_t),
\qquad
q_t=(\Phi_t)_{\#}q_0,
\label{eq:particle_dynamics}
\end{equation}
where $\Phi_t(x)=X_t$ denotes the resulting flow map. When a distribution is represented by a generator, its evolution can be learned by fitting the generator to particle displacements prescribed by this flow. We next describe the energy and velocity used by W-Flow for this purpose.

\subsection{Sinkhorn-Induced Dynamics and One-Step Learning}
\label{subsec:balanced_sinkhorn}
\vspace{-0.5\baselineskip}

W-Flow~\citep{han:2026wflow} instantiates the Wasserstein dynamics in \Secref{subsec:wgf} using the balanced Sinkhorn divergence as its distributional energy. This choice provides a transport-based velocity that can be computed from samples through entropic optimal transport. Although choosing this energy alone does not guarantee convergence of the gradient flow to the target, W-Flow demonstrates that the resulting dynamics can guide one-step generative modeling at scale.

For $q,p\in\mathcal{P}_2(\mathbb{R}^d)$ and the quadratic ground cost $c(x,y)=\frac{1}{2}\|x-y\|_2^2$, the entropically regularized optimal transport cost is
\begin{equation}
\mathrm{OT}_{\varepsilon}(q,p)
=
\inf_{\pi\in\Pi(q,p)}
\left\{
\int c(x,y)\,d\pi(x,y)
+
\varepsilon\,\mathrm{KL}\!\left(\pi\,\|\,q\otimes p\right)
\right\},
\label{eq:balanced_ot}
\end{equation}
where $\Pi(q,p)$ is the set of couplings with marginals $q$ and $p$. Entropic regularization enables computation through Sinkhorn iterations~\citep{cuturi:2013sinkhorn}, but generally gives a nonzero self-cost, $\mathrm{OT}_{\varepsilon}(p,p)\neq 0$. The Sinkhorn divergence~\citep{feydy:2019sinkhorn} removes this self-cost:
\begin{equation}
\mathcal{S}_{\varepsilon}(q,p)
=
\mathrm{OT}_{\varepsilon}(q,p)
-
\frac{1}{2}\mathrm{OT}_{\varepsilon}(q,q)
-
\frac{1}{2}\mathrm{OT}_{\varepsilon}(p,p).
\label{eq:balanced_sinkhorn}
\end{equation}

Let $\pi_{q,p}$ and $\pi_{q,q}$ be optimal couplings for the cross and self transport problems, respectively, and define their barycentric projections as $T_{q,p}(x)=\int y\,\pi_{q,p}(dy\mid x)$ and $T_{q,q}(x)=\int x'\,\pi_{q,q}(dx'\mid x)$. Suppressing the dependence on $\varepsilon$ in this notation, the Wasserstein gradient of $\mathcal{S}_{\varepsilon}(q,p)$ induces the velocity
\begin{equation}
V_{q,p}(x)
=
T_{q,p}(x)-T_{q,q}(x).
\label{eq:wflow_velocity}
\end{equation}
At $q=p$, the cross and self terms coincide, so $V_{p,p}=0$. \citet{han:2026wflow} approximate the continuous particle dynamics in \Eqref{eq:particle_dynamics} with an explicit Euler step. At a current distribution $q$, this gives
\begin{equation}
g_{q,p}(x)
=
x+\eta V_{q,p}(x),
\qquad
q^{+}
=
(g_{q,p})_{\#}q,
\label{eq:particle_update}
\end{equation}
where $\eta>0$ is the step size. To learn these updates with a one-step generator, let $p_{\mathrm{ref}}$ be a reference distribution and $f_{\bar\theta}$ the current generator, inducing $q_{\bar\theta}=(f_{\bar\theta})_{\#}p_{\mathrm{ref}}$. With $\bar\theta$ fixed when constructing the target, \citet{han:2026wflow} minimize
\begin{equation}
\mathcal L_{\mathrm{W\text{-}Flow}}(\theta)
=
\mathbb E_{z\sim p_{\mathrm{ref}}}
\left[
\left\|
f_\theta(z)
-
\operatorname{sg}\!\left[
f_{\bar\theta}(z)
+
\eta V_{q_{\bar\theta},p}
\bigl(f_{\bar\theta}(z)\bigr)
\right]
\right\|_2^2
\right],
\label{eq:wflow_regression_target}
\end{equation}
where $\operatorname{sg}$ denotes stop-gradient. The generator learns the particle updates during training and produces samples in a single forward pass at inference. In practice, the cross and self barycentric terms are estimated from finite minibatches; we examine how this estimation perturbs the population velocity field in \Secref{sec:method}.

\section{\methodname}
\label{sec:method}
\vspace{-0.5\baselineskip}

The balanced Sinkhorn field in \Eqref{eq:wflow_velocity} is defined at the population level, while in practice it is estimated from finite mini-batches~\citep{han:2026wflow}.
We first examine how empirical marginal fluctuations distort the balanced transport geometry (\Secref{sec:method_problem}).
Motivated by this, we introduce \textbf{\methodname}, which relaxes the hard marginal constraints of balanced OT using unbalanced transport penalties (\Secref{sec:method_uot}).
We then analyze how this relaxation affects finite-batch stability relative to balanced OT (\Secref{sec:method_stability}).

\subsection{Finite-Batch Velocity Error in Balanced Sinkhorn Fields}
\label{sec:method_problem}
\vspace{-0.5\baselineskip}

Balanced transport must match empirical marginals exactly, so sampling fluctuations can perturb the barycentric field. To examine this effect, draw three mutually independent batches of $B$ i.i.d.\ samples, $x_1,\ldots,x_B\sim q$, $y_1,\ldots,y_B\sim p$, and $x'_1,\ldots,x'_B\sim q$, and define
\begin{equation}
\widehat q_B=\frac{1}{B}\sum_{i=1}^{B}\delta_{x_i},
\qquad
\widehat p_B=\frac{1}{B}\sum_{j=1}^{B}\delta_{y_j},
\qquad
\widehat q'_B=\frac{1}{B}\sum_{i=1}^{B}\delta_{x'_i}.
\label{eq:empirical_cross_measures}
\end{equation}

The regression targets in \Eqref{eq:wflow_regression_target} use $\widehat V_B:=T_{\widehat q_B,\widehat p_B}-T_{\widehat q_B,\widehat q'_B}$. While prior work studies finite-sample convergence of entropic OT maps~\citep{rigollet:2025sample,goldfeld:2024limit}, we measure the error of this cross-minus-self field at the sampled source points:
\begin{equation}
\mathcal E_{\mathrm{BOT}}(B)
:=
\mathbb E\left[
\frac{1}{B}
\sum_{i=1}^{B}
\left\|
\widehat V_B(x_i)-V_{q,p}(x_i)
\right\|_2^2
\right].
\label{eq:balanced_velocity_mse}
\end{equation}

The following proposition gives a lower bound on finite-batch velocity error under exact marginal matching.

\begin{proposition}[Balanced finite-batch velocity error]
\label{prop:bot_lower_bound}
Assume the cross and self couplings are balanced and the following covariances are finite. For every $B$,
\begin{equation}
\mathcal E_{\mathrm{BOT}}(B)
\ge
\frac{L_{\mathrm{BOT}}^2}{B},
\qquad
L_{\mathrm{BOT}}^2
:=
\operatorname{Tr}\!\left[
\operatorname{Cov}_p(Y)
+
\operatorname{Cov}_q(X)
+
\operatorname{Cov}_q\bigl(V_{q,p}(X)\bigr)
\right].
\label{eq:bot_lower_bound}
\end{equation}
\end{proposition}

The proof is in Appendix~\ref{app:balanced_mse}. Thus, even when the balanced OT field is computed exactly for each minibatch, fluctuations in the empirical means contribute an unavoidable variance component to the velocity error.

\begingroup
\definecolor{velocityPop}{HTML}{708090}
\definecolor{velocityBOT}{HTML}{D94B2B}
\definecolor{velocityUOT}{HTML}{008F87}

\begin{figure}[t]
    \centering
    \captionsetup[subfigure]{font=small,skip=2pt}
    \setlength{\fboxsep}{6pt}
    \setlength{\fboxrule}{0.4pt}

    \begin{subfigure}[t]{0.25\linewidth}
        \vspace{0pt}
        \centering
        \includegraphics[width=\linewidth,trim={40pt 28pt 40pt 28pt},clip]{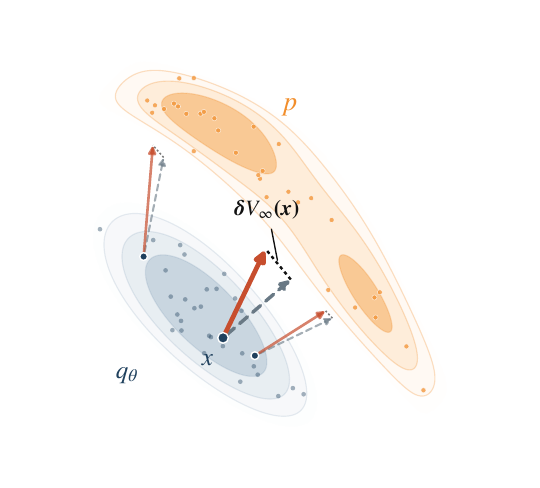}
        \caption{Balanced transport}
        \label{fig:bot_velocity}
    \end{subfigure}%
    \hspace{0.01\linewidth}
    \begin{subfigure}[t]{0.25\linewidth}
        \vspace{0pt}
        \centering
        \includegraphics[width=\linewidth,trim={40pt 28pt 40pt 28pt},clip]{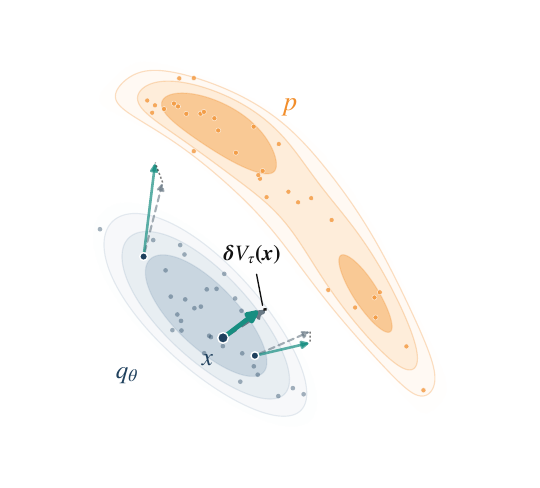}
        \caption{Unbalanced transport}
        \label{fig:uot_velocity}
    \end{subfigure}%
    \hspace{0.01\linewidth}
    \begin{minipage}[t]{0.36\linewidth}
        \vspace{0pt}
        \raggedright
        \fcolorbox{black!35}{white}{%
    \begin{minipage}{\dimexpr\linewidth-2\fboxsep-2\fboxrule\relax}
        \footnotesize
        \begin{tabular*}{\linewidth}{@{}l@{\extracolsep{\fill}}l@{}}
            \tikz[baseline=-0.5ex]\draw[velocityPop,line width=0.9pt,dash pattern=on 3pt off 2pt](0,0)--(0.35,0);~Population
            &
            \tikz[baseline=-0.5ex]\draw[velocityBOT,line width=1pt](0,0)--(0.35,0);~BOT minibatch
            \\
            \tikz[baseline=-0.5ex]\draw[black,line width=0.9pt,densely dotted](0,0)--(0.35,0);~$\delta V$
            &
            \tikz[baseline=-0.5ex]\draw[velocityUOT,line width=1pt](0,0)--(0.35,0);~UOT minibatch
        \end{tabular*}
    \end{minipage}%
}

        \par\vspace{7pt}
\captionsetup{font=small,labelfont=md,justification=raggedright,singlelinecheck=false,skip=0pt}
\caption{\textbf{Finite-batch velocity perturbations.} Both panels use identical samples and arrow scales. Dashed arrows show population velocities, solid arrows show minibatch estimates, and dotted segments mark their difference $\delta V$. UOT has a smaller perturbation at $x$ in this illustration.}
\label{fig:minibatch_uot}
    \end{minipage}
    \par\vspace{-0.8\baselineskip}
\end{figure}
\endgroup

\subsection{\methodname: Relaxing Marginal Constraints}
\label{sec:method_uot}
\vspace{-0.5\baselineskip}

The finite-batch error in \Secref{sec:method_problem} motivates relaxing empirical marginal constraints, allowing transported mass to adapt to sampling fluctuations and potentially reducing velocity perturbations. We therefore use unbalanced optimal transport (UOT)~\citep{chizat:2018uot}, which replaces hard marginal constraints with soft penalties. For $\tau>0$, we use
\begin{equation}
\mathrm{UOT}_{\varepsilon,\tau}(q,p)
=
\inf_{\pi\ge0}
\left\{
\int c(x,y)\,d\pi(x,y)
+
\tau\KL(\pi_1\|q)
+
\tau\KL(\pi_2\|p)
+
\varepsilon\KL\!\left(\pi\,\|\,q\otimes p\right)
\right\},
\label{eq:uot_method}
\end{equation}
where $\pi_1$ and $\pi_2$ are marginals of $\pi$, and $\KL$ denotes the generalized KL divergence. Larger $\tau$ penalizes deviations from $q$ and $p$, while smaller $\tau$ allows trade-offs between marginal mismatch and transport cost. Following \Eqref{eq:balanced_sinkhorn}, we use the debiased unbalanced Sinkhorn divergence~\citep{sejourne:2019sinkhornuot},
\begin{equation}
\mathcal S^{\mathrm U}_{\varepsilon,\tau}(q,p)
=
\mathrm{UOT}_{\varepsilon,\tau}(q,p)
-
\frac{1}{2}\mathrm{UOT}_{\varepsilon,\tau}(q,q)
-
\frac{1}{2}\mathrm{UOT}_{\varepsilon,\tau}(p,p),
\label{eq:uot_sinkhorn_method}
\end{equation}
as our distributional energy, $\mathcal F_\tau(q;p):=\mathcal S^{\mathrm U}_{\varepsilon,\tau}(q,p)$. The mass-correction term in the general unbalanced Sinkhorn divergence vanishes here because $q$ and $p$ both have unit mass.

Let $\pi^{\varepsilon,\tau}_{q,\mu}$ be the optimal UOT coupling between $q$ and $\mu$, where $\mu\in\{p,q\}$. Unlike a balanced coupling, its first marginal need not equal $q$. Writing $\pi^{\varepsilon,\tau}_{q,\mu}(dx,dy)=q(dx)\,\pi^{\varepsilon,\tau}_{q,\mu}(dy\mid x)$, define the local transported-mass ratio and normalized barycentric projection by
\begin{equation}
s^\tau_{q,\mu}(x)
:=
\frac{d(\pi^{\varepsilon,\tau}_{q,\mu})_1}{dq}(x),
\qquad
T^\tau_{q,\mu}(x)
:=
\frac{\int y\,\pi^{\varepsilon,\tau}_{q,\mu}(dy\mid x)}
{s^\tau_{q,\mu}(x)}.
\label{eq:uot_mass_barycenter}
\end{equation}
Here, $\pi^{\varepsilon,\tau}_{q,\mu}(dy\mid x)$ is a possibly unnormalized measure, and the barycenter is defined wherever $s^\tau_{q,\mu}(x)>0$. The Wasserstein gradient of $\mathcal F_\tau$ yields
\begin{equation}
V^\tau_{q,p}(x)
=
s^\tau_{q,p}(x)\bigl(T^\tau_{q,p}(x)-x\bigr)
-
s^\tau_{q,q}(x)\bigl(T^\tau_{q,q}(x)-x\bigr).
\label{eq:uot_wflow_velocity}
\end{equation}

The derivation from the UOT dual and first variation is in Appendix~\ref{app:uot_wgf}. This field retains the cross-minus-self structure in \Eqref{eq:wflow_velocity}, with each displacement weighted by transported mass. Under the assumptions stated in Appendix~\ref{app:identifiability}, it satisfies target identifiability: $V^\tau_{q,p}=0$ $q$-almost everywhere if and only if $q=p$. The proof appears there. Together with marginal relaxation, this property motivates the hypothesis that UOT may improve one-step training when minibatch-induced marginal mismatch perturbs the balanced velocity field. As $\tau\to\infty$, $\mathcal S^{\mathrm U}_{\varepsilon,\tau}(q,p)\to\mathcal S_{\varepsilon}(q,p)$ and $V^\tau_{q,p}\to V_{q,p}$, recovering balanced W-Flow~\citep{han:2026wflow}. The full algorithm is provided in \Secref{app:training_algorithm}.

\subsection{Finite-Batch Velocity Error under Marginal Relaxation}
\label{sec:method_stability}
\vspace{-0.5\baselineskip}

We now compare the finite-batch estimation errors of balanced and unbalanced velocity fields. To this end, define the mass-weighted transport term \(U^\tau_{q,\mu}(x):=s^\tau_{q,\mu}(x)\bigl(T^\tau_{q,\mu}(x)-x\bigr)\). The population field in \Eqref{eq:uot_wflow_velocity} is then \(V^\tau_{q,p}=U^\tau_{q,p}-U^\tau_{q,q}\), and the empirical field is \(\widehat V^\tau_B:=U^\tau_{\widehat q_B,\widehat p_B}-U^\tau_{\widehat q_B,\widehat q'_B}\).

We measure its finite-batch error relative to the population field \(V^\tau_{q,p}\) as
\begin{equation}
\mathcal E_\tau(B)
:=
\mathbb E\left[
\frac{1}{B}
\sum_{i=1}^{B}
\left\|
\widehat V^\tau_B(x_i)-V^\tau_{q,p}(x_i)
\right\|_2^2
\right].
\label{eq:uot_velocity_mse}
\end{equation}
Consistent with the hard-marginal limit \(\tau\to\infty\), we write the balanced quantities from Sec.~\ref{sec:method_problem} as \(\mathcal E_\infty(B):=\mathcal E_{\mathrm{BOT}}(B)\) and \(L_\infty:=L_{\mathrm{BOT}}\). Note that each error is measured relative to its own population field.

\vspace{-0.5\baselineskip}
\paragraph{Finite-batch sensitivity of UOT.}
For the first-order analysis, we assume fixed finite supports with strictly positive population masses, so minibatch sampling induces multinomial fluctuations in the empirical masses. Unlike balanced OT, UOT allows the transported-mass ratio $s_i$ to vary. We analyze how these mass fluctuations affect both $s_i$ and the barycentric destination $T_i$.

\begin{proposition}[Finite-batch error of UW-Flow]
\label{prop:uot_finite_batch_error}
Under the fixed-support assumptions above, for any fixed
$0<\tau<\infty$, as $B\rightarrow \infty$,

\begin{equation}
\mathcal E_\tau(B)
=
\frac{A_\tau}{B}
+
O(B^{-3/2}).
\label{eq:uot_mse_expansion}
\end{equation}
\end{proposition}

The \(1/B\) factor reflects the scale of finite-batch sampling noise, while \(A_\tau<\infty\) is a \(B\)-independent coefficient that quantifies the leading sensitivity of the UOT velocity field to empirical marginal perturbations. The derivation is deferred to Appendix~\ref{app:uot_mse}. Combining this expansion with the balanced lower bound from Sec.~\ref{sec:method_problem} yields the following comparison theorem.

\begin{theorem}[Finite-batch stability comparison]
\label{thm:finite_batch_comparison}
From the finite-batch error characterizations of UOT in \Eqref{eq:uot_mse_expansion} and balanced OT in \Eqref{eq:bot_lower_bound}, if \(A_\tau<L_\infty^2\), then there exists \(B_0<\infty\) such that
\begin{equation}
A_\tau<L_\infty^2
\quad\Longrightarrow\quad
\mathcal E_\tau(B)<\mathcal E_\infty(B),
\qquad \forall B\ge B_0.
\label{eq:uot_bot_comparison}
\end{equation}
\end{theorem}

Under the stated condition, the theorem shows that, for sufficiently large $B$, the empirical UOT field is closer to its own population field than the balanced field is to its population counterpart. Although this comparison does not rank the population fields or guarantee better generation, \Secref{sec:seperate_mode_analysis} establishes the condition in a separated two-mode setting and tests the predicted stability gap with finite-batch experiments. \Secref{sec:results} then reports generation gains in more complex settings.

\subsection{Separate-Mode Analysis}
\label{sec:seperate_mode_analysis}
\vspace{-0.5\baselineskip}

\paragraph{Separated mode analysis.}
To verify the sufficient condition \(A_\tau<L_\infty^2\), we consider a separated two-mode setting. The source and target distributions each consist of two matched modes whose within-pair geometry remains uniformly bounded as the two mode pairs become increasingly separated. Let the target modes have masses \(\omega\) and \(1-\omega\), with conditional means \(\mu_1\) and \(\mu_2\), and define their separation as \(d_\mu:=\|\mu_1-\mu_2\|\).

For balanced OT, the target-mode separation directly contributes to the finite-batch error through the target covariance. In particular, the balanced coefficient satisfies
\begin{equation}
L_\infty^2
\ge
\omega(1-\omega)d_\mu^2.
\label{eq:bot_separated_scaling}
\end{equation}

In contrast, for any fixed finite \(\tau\), UOT can absorb mode-count mismatch through marginal relaxation, so that transport between unmatched modes vanishes as the modes separate while the local transport geometry remains bounded. Consequently,
\begin{corollary}[Bounded UOT sensitivity under mode separation]
\label{cor:uot_separated_scaling}
For any fixed $0<\tau<\infty$, under the separated matched-mode
assumptions above,
\begin{equation}
    A_\tau(d_\mu)=O(1),
    \qquad d_\mu\to\infty.
    \label{eq:uot_separated_scaling}
\end{equation}
\end{corollary}
A formal derivation is provided in Appendix~\ref{app:final_comparison}. Hence, for sufficiently large \(d_\mu\), the condition \(A_\tau<L_\infty^2\) holds.

We use two controlled settings to examine (i) whether the finite-batch error follows the predicted $1/B$ sensitivity, (ii) how their error gap varies with mode separation, (iii) whether the observed advantage extends beyond fixed supports.
We vary $B$, $d$, and $\tau$ in a fixed-support two-particle
model and a continuously sampled two-Gaussian extension.
Experimental details are provided in
Appendix~\ref{app:toy_experiment_setups}.

\vspace{-0.5\baselineskip}
\paragraph{Fixed-support perturbations.}
\Figref{fig:toy-velocity} shows approximately constant
$B\times\mathrm{MSE}$ across the tested batch sizes,
consistent with $1/B$ error scaling for both methods.
The absolute error gap widens with mode separation,
consistent with
\Eqref{eq:bot_separated_scaling} and \Eqref{eq:uot_separated_scaling}.
The $B\times\mathrm{RelMSE}$ curves show that UOT's advantage
at larger separations persists after accounting for
differences in population-field magnitude.
Full results, including cosine similarity, appear in
Appendix~\ref{app:fixed-support-setting}.

\begingroup
\graphicspath{{figures/toy_velocity_main/}{toy_velocity_main/}{./}}

\definecolor{toyBOT}{HTML}{343A40}
\definecolor{toyTauTiny}{HTML}{0072B2}
\definecolor{toyTauSmall}{HTML}{D55E00}
\definecolor{toyTauOne}{HTML}{009E73}
\definecolor{toyTauTen}{HTML}{CC79A7}

\begin{figure*}[t]
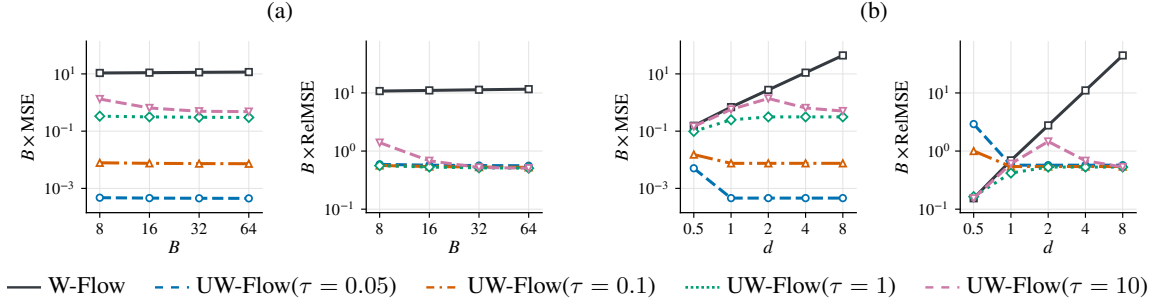

    \centering

    \def\ToyLegendGap{5pt}  %
    \def\ToyTitleGap{2pt}   %
    \def\ToyLabelGap{-5pt}  %

    \par\vspace{\ToyLegendGap}

    \noindent
    \makebox[0.48\linewidth]{\footnotesize (a)}%
    \hspace{0.04\linewidth}%
    \makebox[0.48\linewidth]{\footnotesize (b)}%
    \par\vspace{2pt}
    
    \noindent
    \includegraphics[width=0.235\linewidth]{batch_scaled_mse.pdf}%
    \hspace{0.01\linewidth}%
    \includegraphics[width=0.235\linewidth]{batch_scaled_relmse.pdf}%
    \hspace{0.04\linewidth}%
    \includegraphics[width=0.235\linewidth]{separation_scaled_mse.pdf}%
    \hspace{0.01\linewidth}%
    \includegraphics[width=0.235\linewidth]{separation_scaled_relmse.pdf}%

    \vspace{-0.25\baselineskip}
    
{\footnotesize
\noindent\makebox[\linewidth]{%
    \tikz[baseline=-0.5ex]\draw[toyBOT,line width=0.9pt](0,0)--(0.45,0);~W-Flow
    \hfill
    \tikz[baseline=-0.5ex]\draw[toyTauTiny,line width=0.9pt,dashed](0,0)--(0.45,0);~\methodname{}($\tau=0.05$)
    \hfill
    \tikz[baseline=-0.5ex]\draw[toyTauSmall,line width=0.9pt,dashdotted](0,0)--(0.45,0);~\methodname{}($\tau=0.1$)
    \hfill
    \tikz[baseline=-0.5ex]\draw[toyTauOne,line width=1pt,densely dotted](0,0)--(0.45,0);~\methodname{}($\tau=1$)
    \hfill
    \tikz[baseline=-0.5ex]\draw[toyTauTen,line width=0.9pt,dashed](0,0)--(0.45,0);~\methodname{}($\tau=10$)%
}\par
}

    \vspace{-0.25\baselineskip}

    \caption{\textbf{Minibatch velocity error on fixed supports.}
Each method is evaluated against its own population velocity.
Panel (a) sweeps $B$ at $d=4$, while panel (b) sweeps $d$ at $B=16$.
Within each panel, the left plot shows $B\times\mathrm{MSE}$ and the right plot shows $B\times\mathrm{RelMSE}$.
RelMSE is MSE divided by the corresponding population-field energy.}
    \label{fig:toy-velocity}
    \vspace{-\baselineskip}
\end{figure*}
\endgroup

\vspace{-0.5\baselineskip}
\paragraph{Beyond fixed supports.}
Appendix~\ref{app:gaussian-mix-setting} extends
the comparison to Gaussian mixtures, where both mode
proportions and sample locations fluctuate.
UOT generally exhibits lower relative error and higher
directional agreement over the tested configurations.

\section{Experimental Results}
\label{sec:results}
\vspace{-0.5\baselineskip}

We evaluate \methodname{} on synthetic targets and image-generation benchmarks. Within each comparison, we replace W-Flow's~\citep{han:2026wflow} balanced Sinkhorn energy with its unbalanced counterpart while keeping the generator architecture, optimization protocol, and training budget matched. Each transport group consists of three batches of size $B$: generated samples ($N_{\mathrm{gen}}=B$), target samples ($N_{\mathrm{pos}}=B$), and an independently drawn batch of $B$ generated samples that serves as the self-reference. In transport-batch ablations, we vary $B$ while adjusting the number of groups $N_c$ to keep the number of generated samples per optimizer update fixed at $B_{\mathrm{opt}}=N_cB$.

\vspace{-0.5\baselineskip}

\subsection{Toy Analysis: Synthetic Distribution Matching}
\label{sec:toy_results}
\vspace{-0.5\baselineskip}

We evaluate distribution matching beyond the controlled two-mode settings of \Secref{sec:seperate_mode_analysis}. We train W-Flow and \methodname{} on three two-dimensional targets: checkerboard, two moons, and Gaussian ring. Both methods use $B=128$ and $B_{\mathrm{opt}}=512$, with $\tau=0.1$ for \methodname{}. Experimental details and additional ablations over $B$ and $\tau$ are provided in Appendix~\ref{app:experiments_5.1}.

Qualitatively, \methodname{} captures the characteristic structures of all three targets (\Figref{fig:toy-generation}). The quantitative results in \Tabref{tab:toy-generation} generally favor \methodname{} over W-Flow, as measured by squared sliced Wasserstein distance ($\mathrm{SW}^2$) and squared maximum mean discrepancy ($\mathrm{MMD}^2$).

\noindent
\begin{minipage}{\linewidth}
    \centering
    \providecommand{\ToyAssetPath}{figures/toy_main_layout}
    \graphicspath{{\ToyAssetPath/}}

    \def\ToyQualFraction{0.42}
    \def\ToyTableFraction{0.5}
    \def\ToyTableFont{\footnotesize}
    \def\ToyTableStretch{1.35}
    \def\ToyCaptionGap{6pt}

    \definecolor{toySampleBOT}{HTML}{C96736}
    \definecolor{toySampleUOT}{HTML}{197F83}

    \ifdefined\ToyMainQualBox\else\newsavebox{\ToyMainQualBox}\fi
    \ifdefined\ToyMainBodyHeight\else\newlength{\ToyMainBodyHeight}\fi
    \sbox{\ToyMainQualBox}{%
        \begin{minipage}{\ToyQualFraction\linewidth}
            \begingroup
\centering
\fontsize{8}{9}\selectfont
\setlength{\tabcolsep}{0pt}
\renewcommand{\arraystretch}{1}
\begin{tabular}{@{}c@{\hspace{0.008\linewidth}}c@{\hspace{0.008\linewidth}}c@{\hspace{0.008\linewidth}}c@{}}
    \makebox[0.036\linewidth]{} &
    \makebox[0.313\linewidth]{Checkerboard} &
    \makebox[0.313\linewidth]{Two moons} &
    \makebox[0.313\linewidth]{Gaussian ring} \\[2pt]
    \raisebox{-0.5\height}{\rotatebox{90}{\textcolor{toySampleBOT}{W-Flow}}} &
    \raisebox{-0.5\height}{\includegraphics[width=0.313\linewidth]{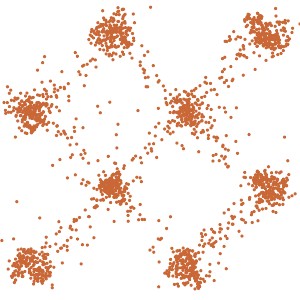}} &
    \raisebox{-0.5\height}{\includegraphics[width=0.313\linewidth]{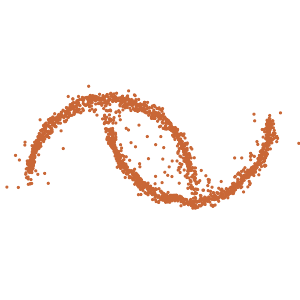}} &
    \raisebox{-0.5\height}{\includegraphics[width=0.313\linewidth]{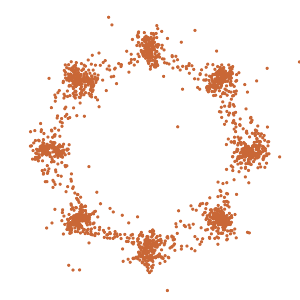}} \\[3pt]
    \raisebox{-0.5\height}{\rotatebox{90}{\textcolor{toySampleUOT}{\methodname{}}}} &
    \raisebox{-0.5\height}{\includegraphics[width=0.313\linewidth]{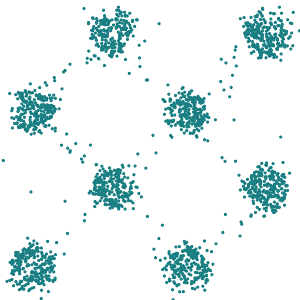}} &
    \raisebox{-0.5\height}{\includegraphics[width=0.313\linewidth]{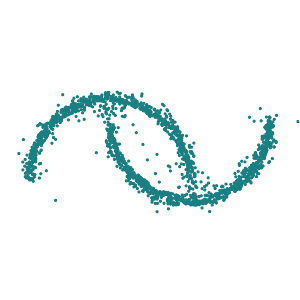}} &
    \raisebox{-0.5\height}{\includegraphics[width=0.313\linewidth]{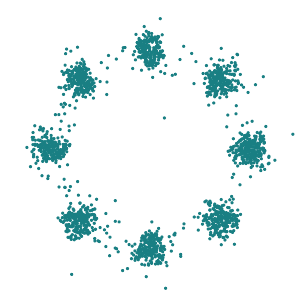}}
\end{tabular}\par
\endgroup

        \end{minipage}%
    }
    \setlength{\ToyMainBodyHeight}{%
        \dimexpr\ht\ToyMainQualBox+\dp\ToyMainQualBox\relax}

    \captionsetup{
        justification=raggedright,
        singlelinecheck=false,
        position=bottom,
        skip=\ToyCaptionGap
    }

\noindent
    \begin{minipage}[t]{\ToyQualFraction\linewidth}
        \vspace{0pt}
        \centering
        \usebox{\ToyMainQualBox}\par
        \captionof{figure}{W-Flow (top) and \methodname{} (bottom) across three distributions.}
        \label{fig:toy-generation}
    \end{minipage}
    \hspace{0.02\linewidth}
    \begin{minipage}[t]{\ToyTableFraction\linewidth}
        \vspace{0.5\baselineskip}
        \captionof{table}{Toy distribution matching with W-Flow and \methodname{}. Values are $\times 10^{-3}$; best in bold.}
        \vspace{-0.5\baselineskip}
        \begin{minipage}[c][\ToyMainBodyHeight][c]{\linewidth}
            \centering

\begingroup
\centering
\footnotesize
\renewcommand{\arraystretch}{1.15}
\begin{tabular*}{\linewidth}{@{}ll|@{\extracolsep{\fill}}ccc@{}}
    \toprule
    & & \multicolumn{3}{c}{Distribution} \\
    \cline{3-5}\addlinespace[2pt]
    Metric & Method & \makecell{Checker-\\board} & \makecell{Two\\moons} & \makecell{Gaussian\\ring} \\
    \midrule
    \multirow{2}{*}{$\mathrm{SW}^{2}\downarrow$}
    & W-Flow       & 2.970 & \textbf{0.825} & 4.280 \\
    & \methodname{} & \textbf{2.600} & 0.852 & \textbf{3.760} \\
    \cmidrule(lr){1-5}
    \multirow{2}{*}{$\mathrm{MMD}^{2}\downarrow$}
    & W-Flow       & 1.490 & 0.483 & 0.866 \\
    & \methodname{} & \textbf{0.700} & \textbf{0.407} & \textbf{0.501} \\
    \bottomrule
\end{tabular*}\par
\vspace{3pt}
\endgroup
        \end{minipage}\par
        \label{tab:toy-generation}
    \end{minipage}
    \par
    \vspace{-\baselineskip}
\end{minipage}

\par

\subsection{Fixed-Budget Minibatch Discrepancy Ablation}
\label{sec:batch_ablation}
\vspace{-0.5\baselineskip}

We examine whether marginal relaxation improves unconditional generation with limited transport support on MNIST~\citep{lecun:1998mnist} and CIFAR-10~\citep{krizhevsky:2009cifar10}. We vary $B\in\{4,8,16,32,64\}$ while keeping $B_{\mathrm{opt}}$ fixed at $64$ for MNIST and $128$ for CIFAR-10. We report Fréchet Inception Distance (FID)~\citep{Heusel:2017FID}, precision, and recall~\citep{kynkaanniemi:2019pr} on both datasets, and Inception Score (IS)~\citep{Salimans:2016IS} on CIFAR-10. Further experimental details are provided in Appendix~\ref{app:experiments}.

\Tabref{tab:mnist_support_ablation} and \Tabref{tab:cifar10_support_ablation} show the generation results as the transport support size $B$ varies under a fixed per-update budget. FID generally decreases with increasing $B$ for both W-Flow and the reported UW-Flow configurations. We evaluate $\tau\in\{0.1,1,10\}$ at each support size and report a selected UW-Flow configuration in the main tables; the complete results are provided in Appendix~\ref{app:batch_ablation}.~\methodname{} consistently achieves lower FID than W-Flow at every tested support size on both datasets. At the smallest support sizes, $B=4$ and $B=8$, UW-Flow achieves the largest absolute FID reductions over W-Flow, by $15.02$ and $14.69$ points, respectively. It also maintains a lower FID at $B=64$ ($4.51$ versus $8.10$). These results show that an appropriate choice of $\tau$ can improve one-step generation by relaxing the hard marginal constraints of balanced OT.

\begingroup
\setlength{\intextsep}{10pt} %

\begin{table}[H]
    \captionsetup{
        skip=3pt,
        justification=raggedright,
        singlelinecheck=false,
        position=top
    }
    \footnotesize
    \setlength{\tabcolsep}{0pt}

    \begin{minipage}[t]{0.45\linewidth}
        {\centering
        \captionsetup{justification=centering}
        \caption{MNIST results with $B_{\mathrm{opt}}=64$.}
        \label{tab:mnist_support_ablation}  
        \begin{tabular*}{\linewidth}{
    @{}c@{\hspace{2pt}}c@{\hspace{2pt}}c@{\hspace{5pt}}l@{\extracolsep{\fill}}ccc@{}
}
            \toprule
            $N_c$ & $N_{\mathrm{gen}}$ & $N_{\mathrm{pos}}$
            & Method & FID $\downarrow$
            & Prec. $\uparrow$ & Rec. $\uparrow$\\
            \midrule

            \multirow{2}{*}{16}
            & \multirow{2}{*}{4}
            & \multirow{2}{*}{4}
            & W-Flow & 19.70 & \textbf{0.54} & 0.06\\
            & & & \methodname
            & \textbf{17.98} & 0.46 & \textbf{0.12}\\
            \cmidrule(lr){1-7}

            \multirow{2}{*}{8}
            & \multirow{2}{*}{8}
            & \multirow{2}{*}{8}
            & W-Flow & 13.69 & 0.46 & 0.24\\
            & & & \methodname
            & \textbf{12.37} & \textbf{0.48} & \textbf{0.28}\\
            \cmidrule(lr){1-7}

            \multirow{2}{*}{4}
            & \multirow{2}{*}{16}
            & \multirow{2}{*}{16}
            & W-Flow & 9.86 & \textbf{0.49} & 0.43\\
            & & & \methodname
            & \textbf{9.22} & 0.47 & \textbf{0.51}\\
            \cmidrule(lr){1-7}

            \multirow{2}{*}{2}
            & \multirow{2}{*}{32}
            & \multirow{2}{*}{32}
            & W-Flow & 8.72 & 0.58 & 0.32\\
            & & & \methodname
            & \textbf{7.33} & \textbf{0.60} & \textbf{0.34}\\
            \cmidrule(lr){1-7}

            \multirow{2}{*}{1}
            & \multirow{2}{*}{64}
            & \multirow{2}{*}{64}
            & W-Flow & 7.03 & 0.57 & 0.53\\
            & & & \methodname
            & \textbf{5.67} & \textbf{0.59} & \textbf{0.57}\\
            \bottomrule
        \end{tabular*}
        }
    \end{minipage}%
    \hspace{0.02\linewidth}%
    \begin{minipage}[t]{0.53\linewidth}
        {\centering
        \captionsetup{justification=centering}
        \caption{CIFAR-10 results with $B_{\mathrm{opt}}=128$.}
        \label{tab:cifar10_support_ablation}
        \begin{tabular*}{\linewidth}{
    @{}c@{\hspace{2pt}}c@{\hspace{2pt}}c@{\hspace{5pt}}l@{\extracolsep{\fill}}cccc@{}
}
            \toprule
            $N_c$ & $N_{\mathrm{gen}}$ & $N_{\mathrm{pos}}$
            & Method & FID $\downarrow$ & IS $\uparrow$
            & Prec. $\uparrow$ & Rec. $\uparrow$\\
            \midrule

            \multirow{2}{*}{32}
            & \multirow{2}{*}{4}
            & \multirow{2}{*}{4}
            & W-Flow & 30.58 & 9.75 & \textbf{0.55} & 0.01\\
            & & & \methodname
            & \textbf{15.56} & \textbf{10.60}
            & 0.52 & \textbf{0.07}\\
            \cmidrule(lr){1-8}

            \multirow{2}{*}{16}
            & \multirow{2}{*}{8}
            & \multirow{2}{*}{8}
            & W-Flow & 24.39 & 8.99 & \textbf{0.58} & 0.01\\
            & & & \methodname
            & \textbf{9.70} & \textbf{10.53}
            & 0.56 & \textbf{0.17}\\
            \cmidrule(lr){1-8}

            \multirow{2}{*}{8}
            & \multirow{2}{*}{16}
            & \multirow{2}{*}{16}
            & W-Flow & 12.49 & \textbf{10.09} & 0.59 & 0.07\\
            & & & \methodname
            & \textbf{10.03} & 9.87
            & \textbf{0.61} & \textbf{0.15}\\
            \cmidrule(lr){1-8}

            \multirow{2}{*}{4}
            & \multirow{2}{*}{32}
            & \multirow{2}{*}{32}
            & W-Flow & 9.57 & 10.32 & \textbf{0.63} & 0.08\\
            & & & \methodname
            & \textbf{5.01} & \textbf{10.43}
            & 0.61 & \textbf{0.39}\\
            \cmidrule(lr){1-8}

            \multirow{2}{*}{2}
            & \multirow{2}{*}{64}
            & \multirow{2}{*}{64}
            & W-Flow & 8.10 & \textbf{10.37} & \textbf{0.60} & 0.27\\
            & & & \methodname
            & \textbf{4.51} & \textbf{10.37}
            & \textbf{0.60} & \textbf{0.45}\\
            \bottomrule
        \end{tabular*}
        }
    \end{minipage}
    \vspace{-\baselineskip}
\end{table}
\endgroup

\Figref{fig:cifar-training} compares FID trajectories across three transport support sizes. UW-Flow with $\tau=0.1$ or $\tau=1$ generally reaches lower FID in fewer training epochs than W-Flow. At $B=4$, UW-Flow with $\tau=0.1$ reaches the FID attained by W-Flow at epoch 500 after approximately 55 epochs, requiring about $9.1\times$ fewer epochs; at $B=64$, it reaches the corresponding FID after 80 epochs, requiring $6.25\times$ fewer epochs. As $\tau$ increases to $10$, the trajectories generally move closer to that of W-Flow.

\begingroup
\graphicspath{{figures/Fig2.CIFAR10_training_curve_logy_panelwise/}{./}}
\definecolor{cifarBOT}{HTML}{343A40}
\definecolor{cifarTauSmall}{HTML}{D55E00}
\definecolor{cifarTauOne}{HTML}{009E73}
\definecolor{cifarTauTen}{HTML}{CC79A7}

\begin{figure*}[t]
    \centering
    \begin{minipage}[c]{0.025\textwidth}
        \centering\rotatebox[origin=c]{90}{\small FID $\downarrow$}
    \end{minipage}\hfill
    \begin{minipage}[c]{0.96\textwidth}
        \begin{minipage}[t]{0.325\linewidth}
            \centering
            {\small (a) $B=4$}\par\vspace{0.3em}
            \includegraphics[width=\linewidth]{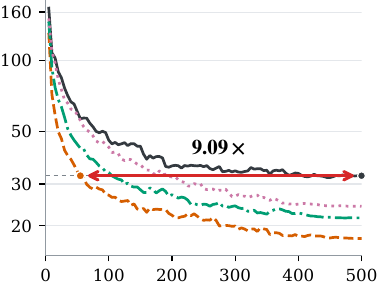}
            \par\vspace{0.1em}{\small Training Epochs}
        \end{minipage}\hfill
        \begin{minipage}[t]{0.325\linewidth}
            \centering
            {\small (b) $B=16$}\par\vspace{0.3em}
            \includegraphics[width=\linewidth]{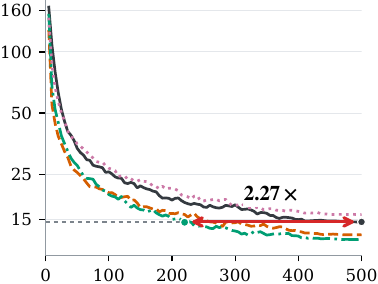}
            \par\vspace{0.1em}{\small Training Epochs}
        \end{minipage}\hfill
        \begin{minipage}[t]{0.325\linewidth}
            \centering
            {\small (c) $B=64$}\par\vspace{0.3em}
            \includegraphics[width=\linewidth]{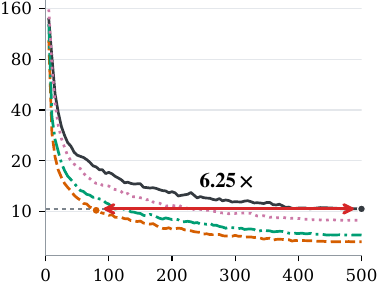}
            \par\vspace{0.1em}{\small Training Epochs}
        \end{minipage}
    \end{minipage}
    \par\vspace{0.65em}
    {\small
    \tikz[baseline=-0.5ex]\draw[cifarBOT,line width=0.9pt] (0,0)--(0.65,0);\ W-Flow
    \quad
    \tikz[baseline=-0.5ex]\draw[cifarTauSmall,line width=0.9pt,dashed] (0,0)--(0.65,0);\ \methodname{}($\tau=0.1$)
    \quad
    \tikz[baseline=-0.5ex]\draw[cifarTauOne,line width=0.9pt,dashdotted] (0,0)--(0.65,0);\ \methodname{}($\tau=1$)
    \quad
    \tikz[baseline=-0.5ex]\draw[cifarTauTen,line width=1pt,densely dotted] (0,0)--(0.65,0);\ \methodname{}($\tau=10$)
    }
    \caption{\textbf{CIFAR-10 training dynamics across transport support sizes.}
    FID versus training epoch for W-Flow and \methodname{} with $\tau\in\{0.1,1,10\}$.
    FID is computed using 10,000 generated samples at each evaluation.
    Panels show $B=4$, $16$, and $64$, with the sample
    budget per optimizer update fixed at
    $B_{\mathrm{opt}}=N_cB=128$.}
    \label{fig:cifar-training}
    \vspace{-1.0\baselineskip}
\end{figure*}

\endgroup

\subsection{Scaling to Complex Visual Distributions}
\label{sec:imagenet100}
\vspace{-0.5\baselineskip}

To evaluate \methodname{} on a more complex target distribution, we use ImageNet-100, a 100-class subset of ImageNet~\citep{deng:2009imagenet}, and test both unconditional and class-conditional generation. We compare W-Flow~\citep{han:2026wflow} and \methodname{} under matched generator architectures, feature encoders, sampling configurations, and optimization budgets. We report FID~\citep{Heusel:2017FID}, IS~\citep{Salimans:2016IS}, precision, and recall~\citep{kynkaanniemi:2019pr}. The experimental setup and results for all tested values of $\tau$ are provided in Appendix~\ref{app:imagenet_100}.

\setlength{\columnsep}{16pt}
\begin{wraptable}{r}{0.50\textwidth}
    \centering
    \vspace{-1.0\baselineskip}
    \captionsetup{
        skip=4pt,
        justification=raggedright,
        singlelinecheck=false
    }
    \captionsetup{justification=centering}
    \caption{ImageNet-100 results at matched budgets.}
    \begingroup
\centering
\footnotesize
\renewcommand{\arraystretch}{1.0}

\begin{tabular*}{\linewidth}{
    @{\extracolsep{\fill}}llcccc@{}
}
    \toprule
    Setting & Method
    & FID $\downarrow$ & IS $\uparrow$
    & Prec. $\uparrow$ & Rec. $\uparrow$ \\
    \midrule
    Uncond & W-Flow
    & 55.25 & 23.89 & \textbf{0.42} & 0.14 \\
    Uncond & \methodname{}
    & \textbf{46.17} & \textbf{27.71}
    & 0.40 & \textbf{0.29} \\
    \specialrule{0.3pt}{2pt}{2pt}
    Cond & W-Flow
    & 9.43 & 60.31 & 0.74 & \textbf{0.26} \\
    Cond & \methodname{}
    & \textbf{9.11} & \textbf{61.34}
    & \textbf{0.75} & \textbf{0.26} \\
    \bottomrule
\end{tabular*}%
\endgroup

    \label{tab:imagenet100}
\end{wraptable}

\Tabref{tab:imagenet100} reports the results for unconditional and class-conditional ImageNet-100 generation. We evaluate $\tau\in\{0.1,1,10\}$ and report the configuration with the lowest FID for each setting. For unconditional generation, UW-Flow with $\tau=0.1$ reduces FID from $55.25$ to $46.17$ ($16.4\%$), with higher IS and recall but lower precision. For class-conditional generation, $\tau=1$ reduces FID from $9.43$ to $9.11$ ($3.4\%$) and improves IS and precision, while recall remains unchanged at the reported precision. The smaller conditional gain is consistent with class-wise transport reducing opportunities for between-class mismatch, although these results do not directly measure coupling mismatch. Full results for all tested values of $\tau$ are provided in Appendix~\ref{app:imagenet_100}.

\FloatBarrier

\section{Conclusion}
\label{sec:conclusion}
\vspace{-0.5\baselineskip}

We introduced UW-Flow to reduce the sensitivity of W-Flow's velocity field to finite-batch marginal fluctuations. Our analysis characterizes the minibatch-to-population error and identifies a separated-mode setting in which marginal relaxation improves finite-batch stability. Controlled experiments support this analysis, while results on MNIST, CIFAR-10, and ImageNet-100 show gains in generation quality under finite transport support.

\vspace{-0.5\baselineskip}
\paragraph{Limitations.} Our stability comparison measures each minibatch field against its own population field. It does not compare the population velocities of UOT and balanced transport or guarantee better generation for arbitrary distributions. We also do not provide a method for selecting $\tau$ for a given setting.

\section*{AI use statement}
\vspace{-0.5\baselineskip}
Generative AI tools were used in this work only for two supporting tasks: revising the manuscript for clarity and readability, and proofreading the mathematical derivations for typographical and notational errors. The research itself was carried out by the authors. This includes the idea of using the debiased unbalanced Sinkhorn divergence as the energy for W-Flow, the derivation of the UW-Flow velocity and its identifiability property, the finite-batch error analysis of balanced and unbalanced transport. The authors were also solely responsible for designing the experiments, implementing the method, selecting datasets, baselines, and metrics, interpreting the results, and surveying and positioning the related literature. AI tools were not used for any of these tasks.

Every AI-suggested edit was checked by the authors, and any change affecting a mathematical expression was verified against the original derivation before being adopted. The authors take full responsibility for the final content of this work, including text, claims or artifacts produced with the aid of generative AI.

\section*{Ethics Statement}
\vspace{-0.5\baselineskip}
This work uses only publicly available datasets (MNIST, CIFAR-10, and ImageNet-100) and synthetic distributions, and it does not involve human-subject experiments or the collection of personally identifiable information. As with other generative models, models trained with our method may inherit biases present in the training data and could be misused to produce misleading synthetic images. We do not identify ethical concerns beyond those generally associated with generative modeling.

\section*{Reproducibility statement}
\vspace{-0.5\baselineskip}
All theoretical results are stated with their assumptions in the main text, and complete proofs are provided in Appendix. Appendix E also describes the setup of the velocity-error experiments, and Appendix F provides the full training configurations, hyperparameters, and complete results for the generation experiments. All datasets used in this work are publicly available, and we will release our code upon acceptance.

\bibliography{iclr2027_conference}
\bibliographystyle{iclr2027_conference}

\appendix
\section{Energy Functional as an Unbalanced Sinkhorn Divergence}
\label{app:uot_energy}

This section establishes the two population-level properties used in
Sec.~\ref{sec:method_uot}.
We first derive the Wasserstein velocity induced by the debiased
unbalanced Sinkhorn divergence and then show that the resulting field
preserves the desired population target.

\subsection{From the UOT Energy to the Wasserstein Velocity}
\label{app:uot_wgf}

For positive measures $\mu\ll\nu$, define the generalized KL divergence
\begin{equation}
    \KL(\mu\|\nu)
    =
    \int
    \log\left(
        \frac{d\mu}{d\nu}
    \right)d\mu
    -
    m(\mu)
    +
    m(\nu).
\end{equation}
Using the convex conjugate of
$h(r)=r\log r-r+1$, namely $h^\star(z)=e^z-1$,
the entropic KL-UOT problem admits the dual
\begin{align}
    \mathrm{UOT}_{\varepsilon,\tau}(q,p)
    =
    \sup_{f,g}
    \Bigg\{
        &
        \tau\int
        \left(
            1-e^{-f(x)/\tau}
        \right)dq(x)
        \nonumber\\
        &+
        \tau\int
        \left(
            1-e^{-g(y)/\tau}
        \right)dp(y)
        \nonumber\\
        &-
        \varepsilon
        \iint
        \left[
            \exp\left(
                \frac{
                    f(x)+g(y)-c(x,y)
                }{\varepsilon}
            \right)-1
        \right]
        dq(x)dp(y)
    \Bigg\}.
    \label{eq:app_uot_dual}
\end{align}

The optimal coupling has Gibbs form
\begin{equation}
    d\pi_{q,p}^{\varepsilon,\tau}(x,y)
    =
    \exp\left(
        \frac{
            f_{q,p}(x)+g_{q,p}(y)-c(x,y)
        }{\varepsilon}
    \right)
    dq(x)dp(y).
    \label{eq:app_uot_gibbs}
\end{equation}

Differentiating the dual objective with respect to $f_{q,p}(x)$ gives
\begin{equation}
    e^{-f_{q,p}(x)/\tau}
    =
    \int
    \exp\left(
        \frac{
            f_{q,p}(x)+g_{q,p}(y)-c(x,y)
        }{\varepsilon}
    \right)
    dp(y).
\end{equation}
The right-hand side is the density of the first coupling marginal with
respect to $q$. Define
\begin{equation}
    s^\tau_{q,p}(x)
    :=
    \frac{
        d(\pi_{q,p}^{\varepsilon,\tau})_1
    }{
        dq
    }(x)
    =
    e^{-f_{q,p}(x)/\tau}.
    \label{eq:app_local_mass}
\end{equation}

Let
\begin{equation}
    T^\tau_{q,p}(x)
    :=
    \frac{
        \int y\,
        \pi_{q,p}^{\varepsilon,\tau}(dy\mid x)
    }{
        s^\tau_{q,p}(x)
    }.
\end{equation}
Thus $s^\tau_{q,p}(x)$ determines how much mass is transported from $x$,
whereas $T^\tau_{q,p}(x)$ determines its normalized barycentric destination.

By the envelope theorem,
\begin{equation}
    \frac{
        \delta
        \mathrm{UOT}_{\varepsilon,\tau}(q,p)
    }{
        \delta q
    }(x)
    =
    (\tau+\varepsilon)
    \left(
        1-s^\tau_{q,p}(x)
    \right).
    \label{eq:app_uot_first_variation}
\end{equation}

For the debiased energy
\begin{equation}
    \mathcal F_\tau(q;p)
    :=
    \mathcal S^{\mathrm U}_{\varepsilon,\tau}(q,p),
\end{equation}
symmetry of the self term gives
\begin{equation}
    \frac{
        \delta\mathcal F_\tau
    }{
        \delta q
    }(q;p)(x)
    =
    (\tau+\varepsilon)
    \left(
        s^\tau_{q,q}(x)
        -
        s^\tau_{q,p}(x)
    \right).
    \label{eq:app_sd_first_variation}
\end{equation}

For
$c(x,y)=\frac12\|x-y\|_2^2$,
differentiating the optimality condition yields
\begin{equation}
    \nabla f_{q,p}(x)
    =
    \frac{\tau}{\tau+\varepsilon}
    \left(
        x-T^\tau_{q,p}(x)
    \right).
\end{equation}
Since
$s^\tau_{q,p}(x)=e^{-f_{q,p}(x)/\tau}$,
\begin{equation}
    \nabla s^\tau_{q,p}(x)
    =
    \frac{
        s^\tau_{q,p}(x)
    }{
        \tau+\varepsilon
    }
    \left(
        T^\tau_{q,p}(x)-x
    \right).
\end{equation}

Therefore
\begin{equation}
    V^\tau_{q,p}(x)
    =
    -\nabla_x
    \frac{
        \delta\mathcal F_\tau
    }{
        \delta q
    }(q;p)(x)
\end{equation}
takes the form
\begin{equation}
    V^\tau_{q,p}(x)
    =
    s^\tau_{q,p}(x)
    \left(
        T^\tau_{q,p}(x)-x
    \right)
    -
    s^\tau_{q,q}(x)
    \left(
        T^\tau_{q,q}(x)-x
    \right).
    \label{eq:app_uot_velocity}
\end{equation}

\vspace{\baselineskip}
\subsection{Population Target Consistency}
\label{app:identifiability}

Let
\begin{equation}
    \phi_q(x)
    :=
    \frac{
        \delta\mathcal F_\tau
    }{
        \delta q
    }(q;p)(x),
    \qquad
    V^\tau_{q,p}(x)
    =
    -\nabla\phi_q(x).
\end{equation}
Assume that $p$ and $q$ are supported on a compact, piecewise-$C^1$,
path-connected set $\mathcal{X}\subset\mathbb{R}^d$, that $q$ has full support on $\mathcal{X}$, $\phi_q$ is $C^1$ on
a neighborhood of $\mathcal X$, and
$\mathcal F_\tau(\cdot;p)$ is convex, nonnegative, and definite on
$\mathcal P(\mathcal X)$.

Suppose first that
\begin{equation}
    V^\tau_{q,p}=0
    \qquad q\text{-a.e.}
\end{equation}
Full support and continuity imply
\begin{equation}
    \nabla\phi_q(x)=0,
    \qquad
    \forall x\in\mathcal X.
\end{equation}
Path connectivity therefore gives
\begin{equation}
    \phi_q(x)=C,
    \qquad
    x\in\mathcal X.
\end{equation}

By convexity, for every probability measure $r$ supported on
$\mathcal X$,
\begin{align}
    \mathcal F_\tau(r;p)
    &\ge
    \mathcal F_\tau(q;p)
    +
    \int
    \phi_q(x)\,d(r-q)(x)
    \\
    &=
    \mathcal F_\tau(q;p),
\end{align}
because $r$ and $q$ have the same total mass.
Hence $q$ is a global minimizer over $\mathcal P(\mathcal X)$.
Since
\begin{equation}
    \mathcal F_\tau(p;p)=0
\end{equation}
and $\mathcal F_\tau$ is nonnegative,
\begin{equation}
    \mathcal F_\tau(q;p)=0.
\end{equation}
Definiteness then implies
\begin{equation}
    q=p.
\end{equation}

Conversely, if $q=p$, the cross and self terms in
~\Eqref{eq:app_uot_velocity} coincide, so
\begin{equation}
    V^\tau_{q,p}=0.
\end{equation}
Therefore
\begin{equation}
    V^\tau_{q,p}=0
    \quad q\text{-a.e.}
    \qquad\Longleftrightarrow\qquad
    q=p.
    \label{eq:app_identifiability}
\end{equation}

\section{Finite-Batch Error of Balanced Sinkhorn Fields}
\label{app:balanced_mse}

This section derives a non-asymptotic lower bound on the velocity
MSE of balanced Sinkhorn fields.
The key observation is that exact marginal preservation determines
the source-averaged velocity through the empirical sample means.
Their sampling fluctuations therefore impose an unavoidable
component of the finite-batch error.

\subsection{A Non-Asymptotic Lower Bound on Velocity MSE}
\label{app:balanced_mse_bound}

\paragraph{Mean preservation under balanced transport.}
Let $\pi^c$ denote the balanced coupling between
$\widehat q_B$ and $\widehat p_B$.
Since each empirical atom has mass $1/B$, its barycentric
projection is
$T_{\widehat q_B,\widehat p_B}(x_i)
=B\sum_j\pi^c_{ij}y_j$.
The exact target marginal constraint,
$\sum_i\pi^c_{ij}=1/B$, then gives
\[
    \frac1B\sum_{i=1}^B
    T_{\widehat q_B,\widehat p_B}(x_i)
    =
    \sum_j y_j\sum_i\pi^c_{ij}
    =
    \frac1B\sum_{j=1}^B y_j
    =
    \overline Y_B.
\]
Applying the same argument to the self coupling gives
$\frac1B\sum_iT_{\widehat q_B,\widehat q'_B}(x_i)
=\overline X'_B$.
Consequently,
\begin{equation}
    \int \widehat V_{\infty,B}(x)\,
    d\widehat q_B(x)
    =
    \overline Y_B-\overline X'_B.
    \label{eq:app_balanced_mean}
\end{equation}
Thus, although individual barycentric assignments depend on the
transport geometry, their source average is fixed by the
empirical marginals.

\paragraph{From the mean error to a lower bound.}
Let
$V_\infty^{\mathrm{pop}}=T_{q,p}-T_{q,q}$
denote the population field, and define
$\Delta_B(x)
:=\widehat V_{\infty,B}(x)-V_\infty^{\mathrm{pop}}(x)$.
Write
$\overline\Delta_B:=\int\Delta_B\,d\widehat q_B$
for its empirical source average.
Since $\widehat q_B$ is a probability measure, Jensen's
inequality yields
\begin{equation}
    \mathcal E_\infty(B)
    :=
    \mathbb E\!\left[
        \|\Delta_B\|_{\widehat q_B}^2
    \right]
    \ge
    \mathbb E\!\left[
        \|\overline\Delta_B\|_2^2
    \right],
    \label{eq:app_balanced_jensen}
\end{equation}
where
$\|\Delta_B\|_{\widehat q_B}^2
=\frac1B\sum_i\|\Delta_B(x_i)\|_2^2$.
It therefore suffices to compute the second moment of the
batch-averaged error.

By~\Eqref{eq:app_balanced_mean},
\begin{equation}
    \overline\Delta_B
    =
    \overline Y_B
    -
    \overline X'_B
    -
    \frac1B\sum_{i=1}^B
    V_\infty^{\mathrm{pop}}(x_i).
    \label{eq:app_balanced_error_mean}
\end{equation}
The same marginal-preservation identity at the population level
gives
\[
    \mathbb E_q\!\left[V_\infty^{\mathrm{pop}}(X)\right]
    =
    \mathbb E_p[Y]-\mathbb E_q[X].
\]
Hence $\mathbb E[\overline\Delta_B]=0$.
Moreover, the three sample averages in~\Eqref{eq:app_balanced_error_mean} depend on mutually
independent batches.
Using the covariance of an i.i.d.\ sample mean, we obtain
\begin{equation}
    \begin{aligned}
        \mathbb E\|\overline\Delta_B\|_2^2
        &=
        \operatorname{Tr}
        \operatorname{Cov}(\overline\Delta_B)\\
        &=
        \frac1B\operatorname{Tr}\!\left[
            \operatorname{Cov}_p(Y)
            +
            \operatorname{Cov}_q(X)
            +
            \operatorname{Cov}_q
            \!\left(V_\infty^{\mathrm{pop}}(X)\right)
        \right]
        =:
        \frac{L_\infty^2}{B}.
    \end{aligned}
    \label{eq:app_balanced_mean_mse}
\end{equation}
Combining this identity with~\Eqref{eq:app_balanced_jensen} proves
\begin{equation}
        {
        \mathcal E_\infty(B)
        \ge
        \frac{L_\infty^2}{B}
    },
    \qquad B\ge1.
    \label{eq:app_balanced_lower}
\end{equation}

This bound captures the component of velocity error that remains
after averaging over source particles.
It follows from exact marginal preservation and batch independence.

\section{Finite-Batch Error of UW-Flow}
\label{app:uot_mse}

This section derives the finite-batch velocity MSE of UW-Flow.
Because UOT does not preserve the input marginals exactly, the
mean-preservation argument in Appendix~\ref{app:balanced_mse_bound}
does not directly apply.
Instead, we analyze how empirical mass fluctuations perturb the
mass factors and barycentric projections that define the field.
We first derive the linear response of a single UOT field, then
combine the cross and self responses to obtain the velocity MSE,
and finally bound its leading coefficient.

Throughout, we fix $\varepsilon>0$ and $0<\tau<\infty$ and consider
probability measures on fixed finite supports:
\begin{equation}
    q=\sum_{i=1}^{M_q}a_i\delta_{x_i},
    \qquad
    p=\sum_{j=1}^{M_p}b_j\delta_{y_j},
    \qquad
    a_i,b_j\ge m_0>0,
    \label{eq:app_uot_support}
\end{equation}
where $\sum_i a_i=\sum_j b_j=1$.
We use the quadratic cost
$C_{ij}=\frac12\|x_i-y_j\|_2^2$.
For a nonnegative weight vector $\nu$, write
$\|z\|_\nu^2:=\sum_i\nu_i\|z_i\|_2^2$;
unsubscripted norms are Euclidean norms of the stacked vectors.

\subsection{First-Order Perturbation of the UOT Field}
\label{app:uot_linear_response}

\paragraph{Population coupling and conditional transport.}
The population KL-UOT coupling has the form
\begin{equation}
    \pi_{ij}=a_i u_iK_{ij}b_jv_j,
    \qquad
    K_{ij}=e^{-C_{ij}/\varepsilon},
    \qquad
    \lambda=\frac{\tau}{\tau+\varepsilon}\in(0,1),
\end{equation}
where the generalized Sinkhorn scalings satisfy
\begin{equation}
    u=[K(b\odot v)]^{-\lambda},
    \qquad
    v=[K^\top(a\odot u)]^{-\lambda}.
    \label{eq:app_uot_scalings}
\end{equation}
All powers, logarithms, and products denoted by $\odot$ are
entrywise.

Let $m^\pi=\pi\mathbf1$ and $n^\pi=\pi^\top\mathbf1$ be the
transported marginals, and define the conditional coupling matrices
\begin{equation}
    P_{ij}=\frac{\pi_{ij}}{m_i^\pi},
    \qquad
    Q_{ji}=\frac{\pi_{ij}}{n_j^\pi}.
    \label{eq:app_uot_conditionals}
\end{equation}
Thus, $P$ averages target quantities conditional on a source atom,
while $Q$ averages source quantities conditional on a target atom.

The identity
$m_i^\pi P_{ij}=\pi_{ij}=n_j^\pi Q_{ji}$
shows that $Q=P^\ast$ in the weighted spaces
$L^2(m^\pi)$ and $L^2(n^\pi)$.
Conditional Jensen's inequality also makes both operators
contractions. Consequently,
\begin{equation}
    0\preceq QP\preceq I,
    \qquad
    \bigl\|(I-\lambda^2QP)^{-1}\bigr\|_{n^\pi\to n^\pi}
    \le \frac{1}{1-\lambda^2}.
    \label{eq:app_uot_resolvent}
\end{equation}

\paragraph{Response of the Sinkhorn scalings.}
We perturb the source and target masses while keeping their
locations fixed.
Let
\[
    r=\log u,\qquad t=\log v,\qquad
    D_a=\operatorname{diag}(a),\qquad
    D_b=\operatorname{diag}(b),
\]
and define the relative perturbations
\begin{equation}
    \rho_a=D_a^{-1}\delta a,
    \qquad
    \rho_b=D_b^{-1}\delta b.
\end{equation}
Differentiating the logarithmic form of
\Eqref{eq:app_uot_scalings} gives
\begin{equation}
    \delta r=-\lambda P(\rho_b+\delta t),
    \qquad
    \delta t=-\lambda Q(\rho_a+\delta r).
\end{equation}
The quantity
$w:=\rho_b+\delta t=\delta\log(b\odot v)$
collects the change in the effective target weights.
Eliminating $\delta r$ yields
\begin{equation}
    w
    =
    (I-\lambda^2QP)^{-1}
    (\rho_b-\lambda Q\rho_a).
    \label{eq:app_uot_w}
\end{equation}
The inverse is well defined by \Eqref{eq:app_uot_resolvent}.

\paragraph{Response of the transport field.}
Define the transported-mass factor and barycentric projection by
\begin{equation}
    s_i=\frac{m_i^\pi}{a_i},
    \qquad
    T_i=\sum_jP_{ij}y_j,
    \qquad
    U_{\tau,i}=s_i(T_i-x_i).
\end{equation}
Marginal perturbations affect both the conditional destination
$T_i$ and the mass factor $s_i$.
Their first-order changes are
\begin{equation}
    \begin{aligned}
        \delta P_{ij}
        &=P_{ij}\bigl[w_j-(Pw)_i\bigr],\\
        \delta T_i
        &=\sum_jP_{ij}(y_j-T_i)w_j,\\
        \delta s_i
        &=s_i(1-\lambda)(Pw)_i.
    \end{aligned}
    \label{eq:app_uot_component_responses}
\end{equation}
For the last identity, we used
$s_i=[K(b\odot v)]_i^{1-\lambda}$.
Applying the product rule to $U_{\tau,i}=s_i(T_i-x_i)$ gives
\begin{equation}
    \delta U_{\tau,i}
    =
    s_i\left[
        \sum_jP_{ij}(y_j-T_i)w_j
        +(1-\lambda)(T_i-x_i)(Pw)_i
    \right].
    \label{eq:app_uot_deltaU}
\end{equation}
Let $H$ denote this linear map, so that $\delta U_\tau=Hw$.
The two terms in \Eqref{eq:app_uot_deltaU} account for changes in
the barycentric projection and the transported-mass factor,
respectively.

\subsection{Finite-Batch Velocity MSE}
\label{app:uot_velocity_mse}

\paragraph{Sampling fluctuations and the debiased response.}
Under i.i.d.\ sampling from \Eqref{eq:app_uot_support}, the empirical
mass vectors satisfy
\begin{equation}
    B\widehat a_B\sim\operatorname{Multinomial}(B,a),
    \qquad
    B\widehat b_B\sim\operatorname{Multinomial}(B,b),
    \qquad
    B\widehat a'_B\sim\operatorname{Multinomial}(B,a),
\end{equation}
independently.
These vectors define $\widehat q_B$, $\widehat p_B$, and
$\widehat q'_B$ on the fixed supports.
The population and minibatch fields are
\begin{equation}
    \begin{aligned}
        V_\tau^{\mathrm{pop}}
        &=U_\tau(q,p)-U_\tau(q,q),\\
        \widehat V_{\tau,B}
        &=U_\tau(\widehat q_B,\widehat p_B)
          -U_\tau(\widehat q_B,\widehat q'_B).
    \end{aligned}
\end{equation}

To separate the batch-size dependence from the field response,
define
\begin{equation}
    \begin{aligned}
        \xi_a&=\sqrt B\,D_a^{-1}(\widehat a_B-a),\\
        \xi_b&=\sqrt B\,D_b^{-1}(\widehat b_B-b),\\
        \xi'_a&=\sqrt B\,D_a^{-1}(\widehat a'_B-a).
    \end{aligned}
\end{equation}
These vectors are centered and mutually independent, with
\begin{equation}
    \operatorname{Cov}(\xi_a)
    =
    \operatorname{Cov}(\xi'_a)
    =
    D_a^{-1}-\mathbf1\mathbf1^\top,
    \qquad
    \operatorname{Cov}(\xi_b)
    =
    D_b^{-1}-\mathbf1\mathbf1^\top.
    \label{eq:app_uot_noise_covariance}
\end{equation}
In particular, their covariances are independent of $B$.

Use subscripts $c$ and $s$ for quantities associated with the
population cross coupling $(q,p)$ and self coupling $(q,q)$,
respectively.
Applying \Eqref{eq:app_uot_w} to the scaled fluctuations gives
\begin{equation}
    \begin{aligned}
        w_c
        &=
        (I-\lambda^2Q_cP_c)^{-1}
        (\xi_b-\lambda Q_c\xi_a),\\
        w_s
        &=
        (I-\lambda^2P_s^2)^{-1}
        (\xi'_a-\lambda P_s\xi_a),
    \end{aligned}
\end{equation}
where $P_s=Q_s$ follows from symmetry of the population self coupling.
The scaled first-order velocity response and its second moment are
\begin{equation}
    L_\tau=H_cw_c-H_sw_s,
    \qquad
    A_\tau:=\mathbb E\|L_\tau\|_a^2.
    \label{eq:app_uot_Atau}
\end{equation}
The cross and self responses share the source fluctuation $\xi_a$;
their dependence is therefore retained in $A_\tau$.
Since $L_\tau$ is linear in noises with the covariances in
\Eqref{eq:app_uot_noise_covariance}, $A_\tau$ is independent of $B$.

\paragraph{From linear response to expected MSE.}
Let
\[
    \delta\theta_B
    :=
    (\widehat a_B-a,\widehat b_B-b,\widehat a'_B-a),
    \qquad
    e_B:=\widehat V_{\tau,B}-V_\tau^{\mathrm{pop}}.
\]
Multinomial sampling gives
\begin{equation}
    \mathbb E\|\delta\theta_B\|^2=O(B^{-1}),
    \qquad
    \mathbb E\|\delta\theta_B\|^4=O(B^{-2}).
    \label{eq:app_uot_mass_moments}
\end{equation}

The simultaneous log-scaling map is a $\lambda$-contraction in
the sup norm and is smooth for positive input masses.
Its fixed point, and hence the velocity map, is therefore smooth
in the interior of the probability simplices.
On the event that every empirical mass is at least $m_0/2$,
a second-order Taylor expansion yields
\begin{equation}
    e_B=\frac1{\sqrt B}L_\tau+R_B,
    \qquad
    \|R_B\|\le C\|\delta\theta_B\|^2.
    \label{eq:app_uot_taylor}
\end{equation}

We also account for batches with small or zero counts.
The probability of the complementary event is at most
$C_0e^{-cB}$.
Moreover, the strictly positive kernels on fixed finite supports
give a bounded extension of the field to all probability weights.
Indeed, for input weights $(\alpha,\beta)$, set
\[
    z_i=[K(\beta\odot v)]_i,\qquad
    s_i=z_i^{1-\lambda},\qquad
    P_{ij}=\frac{K_{ij}\beta_jv_j}{z_i}.
\]
These expressions define $U_i=s_i(T_i-x_i)$ even when $\alpha_i=0$.
The log-scaling contraction bounds the scalings uniformly over
$(\alpha,\beta)$ for the fixed kernel, so this extension is bounded.
Its values at unobserved source atoms do not affect the empirical
norm.

Combining this bounded extension with
\Eqref{eq:app_uot_mass_moments}, the Taylor bound, and local
Lipschitz continuity gives
\begin{equation}
    \mathbb E\|R_B\|^2=O(B^{-2}),
    \qquad
    \mathbb E\|e_B\|^4=O(B^{-2}).
\end{equation}
Consequently, Cauchy--Schwarz yields
\begin{equation}
    \begin{aligned}
        \mathbb E\|e_B\|_a^2
        &=
        \frac{A_\tau}{B}
        +\frac{2}{\sqrt B}
          \mathbb E\langle L_\tau,R_B\rangle_a
        +\mathbb E\|R_B\|_a^2\\
        &=
        \frac{A_\tau}{B}+O(B^{-3/2}).
    \end{aligned}
    \label{eq:app_uot_population_norm_mse}
\end{equation}
Finally, replacing the population weights by the empirical
weights changes the expectation by at most
\begin{equation}
    \begin{aligned}
        \mathbb E\left|
            \|e_B\|_{\widehat a_B}^2-\|e_B\|_a^2
        \right|
        &\le
        \bigl(\mathbb E\|\widehat a_B-a\|_\infty^2\bigr)^{1/2}
        \bigl(\mathbb E\|e_B\|^4\bigr)^{1/2}\\
        &=O(B^{-3/2}).
    \end{aligned}
\end{equation}
Thus,
\begin{equation}
    {
        \mathcal E_\tau(B)
        :=
        \mathbb E\|e_B\|_{\widehat a_B}^2
        =
        \frac{A_\tau}{B}+O(B^{-3/2}).
    }
    \label{eq:app_uot_mse}
\end{equation}
The coefficient $A_\tau$ measures the leading velocity response
to sampling fluctuations, while $1/B$ captures their batch-size
dependence.

\subsection{An Upper Bound on UOT Sensitivity}
\label{app:uot_sensitivity_bound}

We now bound $A_\tau$ through the conditional transport distances
\begin{equation}
    H_{\tau,c}
    :=
    \max_i\sum_jP^c_{ij}\|y_j-x_i\|_2^2,
    \qquad
    H_{\tau,s}
    :=
    \max_i\sum_kP^s_{ik}\|x_k-x_i\|_2^2.
    \label{eq:app_uot_geometry_moments}
\end{equation}
These quantities measure the squared transport distances seen
from each source atom.
We first establish a bound for one coupling and then combine
the cross and self terms.

\paragraph{Controlling the mass and noise response.}
For a population coupling, let
$M_\pi:=\sum_{ij}\pi_{ij}$ and
$\overline s:=\max_i m_i^\pi/a_i$.
The nonnegative cost and probability input masses imply
$M_\pi\le1$.
To see this, write $\pi=M_\pi\gamma$ with
$\sum_{ij}\gamma_{ij}=1$.
The KL-UOT objective takes the form
\[
    M_\pi G(\gamma)
    +(\varepsilon+2\tau)
      (M_\pi\log M_\pi-M_\pi+1),
\]
where $G(\gamma)\ge0$ is the cost-plus-KL objective evaluated at
the normalized plan $\gamma$.
Optimality in $M_\pi$ therefore gives
\begin{equation}
    M_\pi
    =
    e^{-G(\gamma)/(\varepsilon+2\tau)}
    \le1,
    \qquad
    \overline s\le\frac{M_\pi}{m_0}\le\frac1{m_0}.
\end{equation}

For the scaled relative noises,
\Eqref{eq:app_uot_noise_covariance} implies
\[
    \mathbb E\|\xi_a\|_{m^\pi}^2\le\frac{M_\pi}{m_0},
    \qquad
    \mathbb E\|\xi_b\|_{n^\pi}^2\le\frac{M_\pi}{m_0}.
\]
Using the resolvent bound in \Eqref{eq:app_uot_resolvent}
and the contraction property of $Q$, we obtain
\begin{equation}
    \begin{aligned}
        \bigl(\mathbb E\|w\|_{n^\pi}^2\bigr)^{1/2}
        &\le
        \frac{
            \bigl(\mathbb E\|\xi_b\|_{n^\pi}^2\bigr)^{1/2}
            +\lambda
            \bigl(\mathbb E\|\xi_a\|_{m^\pi}^2\bigr)^{1/2}
        }{1-\lambda^2}\\
        &\le
        \frac{\sqrt{M_\pi/m_0}}{1-\lambda}.
    \end{aligned}
    \label{eq:app_uot_noise_bound}
\end{equation}

\paragraph{Controlling the velocity response.}
For a single coupling, write
$H_{\mathrm{geom}}:=\max_i\sum_jP_{ij}\|y_j-x_i\|_2^2$.
Let $d_{ij}:=y_j-x_i$ and
$\overline d_i:=\sum_jP_{ij}d_{ij}=T_i-x_i$.
Then \Eqref{eq:app_uot_deltaU} becomes
\[
    (Hw)_i
    =
    s_i\sum_jP_{ij}(d_{ij}-\lambda\overline d_i)w_j.
\]
Since
\[
    \begin{aligned}
        \sum_jP_{ij}
        \|d_{ij}-\lambda\overline d_i\|_2^2
        &=
        \sum_jP_{ij}\|d_{ij}\|_2^2
        -\lambda(2-\lambda)\|\overline d_i\|_2^2\\
        &\le H_{\mathrm{geom}},
    \end{aligned}
\]
conditional Cauchy--Schwarz gives
\begin{equation}
    \begin{aligned}
        \|Hw\|_a^2
        &\le
        H_{\mathrm{geom}}
        \sum_i a_i s_i^2\sum_jP_{ij}w_j^2\\
        &\le
        \overline s\,H_{\mathrm{geom}}
        \|w\|_{n^\pi}^2.
    \end{aligned}
\end{equation}
Combining this with \Eqref{eq:app_uot_noise_bound} yields
\begin{equation}
    \mathbb E\|Hw\|_a^2
    \le
    \frac{H_{\mathrm{geom}}}{m_0^2(1-\lambda)^2}.
\end{equation}
Applying this bound to the cross and self couplings and using
Minkowski's inequality gives
\begin{equation}
    {
        A_\tau
        \le
        \left(
            \frac{
                \sqrt{H_{\tau,c}}+\sqrt{H_{\tau,s}}
            }{
                m_0(1-\lambda)
            }
        \right)^2.
    }
    \label{eq:app_Atau_bound}
\end{equation}
This step does not require independence between the cross
and self responses.

\paragraph{Bounded sensitivity under mode separation.}
Consider a fixed finite collection of matched source and target
blocks with centers $c_k(d)$:
\begin{equation}
    x_i=c_{k(i)}(d)+\widetilde x_i,
    \qquad
    y_j=c_{\ell(j)}(d)+\widetilde y_j,
    \qquad
    \|\widetilde x_i\|_2,\|\widetilde y_j\|_2\le R_0.
    \label{eq:app_uot_paired_modes}
\end{equation}
Distinct centers are at least $d$ apart, and every block contains
both source and target atoms.
Assume that all within-block cross and self distances are bounded
by $R_{\mathrm{loc}}$.
The atom masses, block structure, $R_0$, and $R_{\mathrm{loc}}$
remain fixed as $d$ increases.

Define the distance-weighted off-block leakage by
\begin{equation}
    \ell_c(d)
    :=
    \max_i
    \sum_{j:\,\ell(j)\ne k(i)}
    P^c_{ij}\|y_j-x_i\|_2^2,
\end{equation}
and define $\ell_s(d)$ analogously for the self coupling.
Then
\begin{equation}
    H_{\tau,c}\le R_{\mathrm{loc}}^2+\ell_c(d),
    \qquad
    H_{\tau,s}\le R_{\mathrm{loc}}^2+\ell_s(d).
    \label{eq:app_uot_local_leakage}
\end{equation}

It remains to control the scalings, since decay of the Gibbs
kernel alone does not imply decay of the conditional coupling.
Set $k_{\mathrm{loc}}:=e^{-R_{\mathrm{loc}}^2/(2\varepsilon)}$.
The existence of a local partner of mass at least $m_0$ gives
\[
    m_0k_{\mathrm{loc}}\le(Kb)_i\le1,
    \qquad
    m_0k_{\mathrm{loc}}\le(K^\top a)_j\le1.
\]
The $\lambda$-contraction of the log-scaling map therefore bounds
its fixed point by
\begin{equation}
    \|(\log u,\log v)\|_\infty
    \le
    \frac{\lambda}{1-\lambda}
    \left(
        \log\frac1{m_0}
        +\frac{R_{\mathrm{loc}}^2}{2\varepsilon}
    \right)
    =:L_\star,
\end{equation}
independently of $d$.
Consequently,
\begin{equation}
    P^c_{ij}
    =
    \frac{K_{ij}b_jv_j}{\sum_\ell K_{i\ell}b_\ell v_\ell}
    \le
    C_\star b_j
    e^{-\|x_i-y_j\|_2^2/(2\varepsilon)},
    \qquad
    C_\star:=\frac{e^{2L_\star}}{m_0k_{\mathrm{loc}}}.
\end{equation}

For off-block pairs, $\|x_i-y_j\|_2\ge d-2R_0$.
Since $r^2e^{-r^2/(2\varepsilon)}$ decreases for
$r\ge\sqrt{2\varepsilon}$, sufficiently large $d$ gives
\begin{equation}
    \ell_c(d)
    \le
    C_\star(d-2R_0)^2
    e^{-(d-2R_0)^2/(2\varepsilon)}
    \longrightarrow0.
\end{equation}
The same argument applies to the self coupling, so
$\ell_s(d)\to0$.
Combining \Eqref{eq:app_uot_local_leakage} with
\Eqref{eq:app_Atau_bound} proves
\begin{equation}
    A_\tau(d)
    \le
    \frac{
        4R_{\mathrm{loc}}^2+2\ell_c(d)+2\ell_s(d)
    }{
        m_0^2(1-\lambda)^2
    }
    =
    O(1),
    \qquad d\to\infty.
    \label{eq:app_Atau_O1}
\end{equation}
Thus, for fixed finite $\tau$ and $\varepsilon$, the leading
UOT sensitivity remains bounded as matched modes separate.

\section{Comparison of Balanced and Unbalanced Finite-Batch Errors}
\label{app:final_comparison}

This section directly compares the finite-batch velocity errors derived
for balanced Sinkhorn transport in Appendix~\ref{app:balanced_mse}
and UW-Flow in Appendix~\ref{app:uot_mse}.
The comparison shows that both errors operate at the same leading
$1/B$ sampling scale, while marginal relaxation can reduce the
corresponding sensitivity coefficient.

\paragraph{Comparison of the leading finite-batch errors.}

Appendix~\ref{app:balanced_mse} gives the non-asymptotic lower bound
\begin{equation}
    \mathcal E_\infty(B)
    \ge
    \frac{L_\infty^2}{B},
    \qquad B\ge1,
    \label{eq:app_balanced_comparison}
\end{equation}
where $L_\infty^2$ is the variance contribution imposed by exact
marginal preservation.

For finite $\tau$, Appendix~\ref{app:uot_mse} gives
\begin{equation}
    \mathcal E_\tau(B)
    =
    \frac{A_\tau}{B}
    +
    O(B^{-3/2}),
    \label{eq:app_uot_comparison}
\end{equation}
where $A_\tau$ is the leading sensitivity of the UOT velocity field
to empirical marginal perturbations.

Suppose
\begin{equation}
    A_\tau < L_\infty^2.
    \label{eq:app_coeff_condition}
\end{equation}
Let
\begin{equation}
    g := L_\infty^2-A_\tau > 0.
\end{equation}
Multiplying~\Eqref{eq:app_uot_comparison} by $B$ gives
\begin{equation}
    B\mathcal E_\tau(B)
    =
    A_\tau
    +
    O(B^{-1/2}).
\end{equation}
Hence, there exists $B_0<\infty$ such that
\begin{equation}
    \left|
        B\mathcal E_\tau(B)-A_\tau
    \right|
    <
    \frac{g}{2},
    \qquad B\ge B_0.
\end{equation}
For such $B$,
\begin{equation}
    B\mathcal E_\tau(B)
    <
    A_\tau+\frac{g}{2}
    <
    L_\infty^2
    \le
    B\mathcal E_\infty(B),
\end{equation}
and therefore
\begin{equation}
    \mathcal E_\tau(B)
    <
    \mathcal E_\infty(B),
    \qquad B\ge B_0.
    \label{eq:app_final_comparison}
\end{equation}

Thus, the comparison does not rely on a different convergence rate
in $B$: both errors scale at the leading $1/B$ order.
The advantage arises when marginal relaxation reduces the leading
finite-batch sensitivity from the balanced level.

\paragraph{Implication under separated modes.}

We now specialize the coefficient comparison to the paired-mode
setting considered in Sec.~\ref{sec:seperate_mode_analysis}.
For any fixed finite $\tau$, Appendix~\ref{app:uot_mse} shows that
\begin{equation}
    A_\tau(d)=O(1)
    \qquad \text{as } d\to\infty,
\end{equation}
provided the local source--target geometry remains bounded.

In contrast, Appendix~\ref{app:balanced_mse} gives
\begin{equation}
    L_\infty^2
    \ge
    \omega(1-\omega)d_\mu^2.
\end{equation}
Therefore, if the separation of the target modes satisfies
\begin{equation}
    d_\mu=\Omega(d),
\end{equation}
then
\begin{equation}
    L_\infty^2(d)=\Omega(d^2).
\end{equation}

Consequently, for sufficiently large mode separation,
$L_\infty^2(d)$ exceeds the bounded UOT coefficient $A_\tau(d)$,
so the condition in~\Eqref{eq:app_coeff_condition} is satisfied.
It follows that
\begin{equation}
    \exists d_0\;
    \forall d\ge d_0\;
    \exists B_0(d)\;
    \forall B\ge B_0(d):
    \qquad
    \mathcal E_\tau(B)
    <
    \mathcal E_\infty(B).
\end{equation}

Hence, in the separated-mode regime, exact marginal preservation can
make the balanced field increasingly sensitive to finite-batch mass
fluctuations, while finite-$\tau$ UOT can keep the leading sensitivity
controlled by the local transport geometry.

\section{Additional Velocity-Field Experiments}
\label{app:toy_experiment_setups}

\subsection{Evaluation Protocol and Metrics}
\label{sec4_4-setup}
\paragraph{Metrics.}
For both experiments, we compare the minibatch velocity
$\widehat V_{\tau,B}$ with its corresponding population field
$V_{\tau}^{\mathrm{pop}}$ using
\[
B\,\mathrm{MSE},
\qquad
\mathrm{RelMSE},
\qquad
\mathrm{Cosine\ Similarity}.
\]

We report the batch-scaled velocity error
\[
B\,\mathrm{MSE}_{\tau}(B)
=
B\left\|
\widehat V_{\tau,B}
-
V_{\tau}^{\mathrm{pop}}
\right\|_{\widehat q_B}^2,
\]
which factors out the leading $1/B$ sampling dependence and facilitates
comparison of finite-batch sensitivity across batch sizes.

Since a smaller population velocity can also reduce the absolute error, we
report the relative MSE
\[
\mathrm{RelMSE}_{\tau}(B)
=
\frac{
\left\|
\widehat V_{\tau,B}
-
V_{\tau}^{\mathrm{pop}}
\right\|_{\widehat q_B}^2
}{
\left\|
V_{\tau}^{\mathrm{pop}}
\right\|_{\widehat q_B}^2
},
\]
which measures the error relative to the population-field magnitude.
Finally, cosine similarity measures directional agreement between the
minibatch and population velocity fields independently of their scale.

\subsection{Fixed-Support Setting}
\label{app:fixed-support-setting}
We consider two matched source--target pairs with fixed supports \[ x_{\pm}=(\pm d/2,0), \qquad y_{\pm}=(\pm d/2,1), \] and assign probability mass $1/2$ to each point. The support locations are fixed across trials, while independent minibatches $\widehat q_B$, $\widehat p_B$, and $\widehat q'_B$ induce fluctuations only in the empirical masses. This directly matches the fixed-support multinomial perturbation setting analyzed in \Secref{sec:method_stability}.

\begin{figure*}[t]
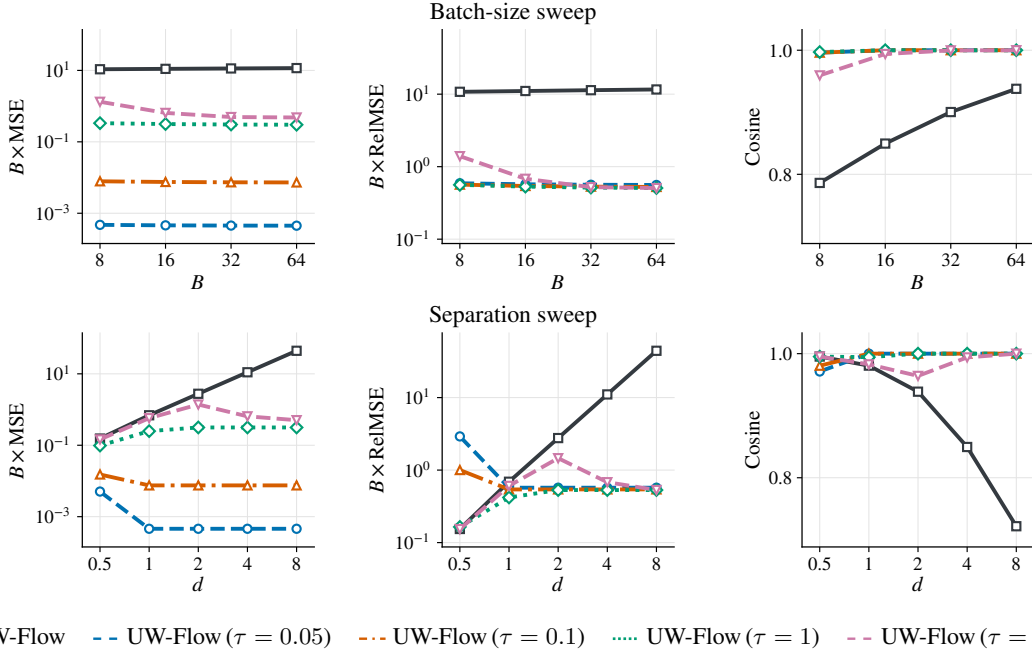

    \centering

    \graphicspath{{figures/toy_velocity_appendix_fixed/}{toy_velocity_appendix_fixed/}{./}}

    \definecolor{toyBOT}{HTML}{343A40}
    \definecolor{toyTauTiny}{HTML}{0072B2}
    \definecolor{toyTauSmall}{HTML}{D55E00}
    \definecolor{toyTauOne}{HTML}{009E73}
    \definecolor{toyTauTen}{HTML}{CC79A7}

    {\small Batch-size sweep\par}
    \vspace{1pt}

    \noindent
    \includegraphics[
        width=0.31\linewidth,
        trim={0mm 2mm 0mm 2mm},
        clip
    ]{batch_scaled_mse.pdf}%
    \hspace{0.005\linewidth}%
    \includegraphics[
        width=0.31\linewidth,
        trim={0mm 2mm 0mm 2mm},
        clip
    ]{batch_scaled_relmse.pdf}%
    \hspace{0.005\linewidth}%
    \includegraphics[
        width=0.31\linewidth,
        trim={0mm 2mm 0mm 2mm},
        clip
    ]{batch_cosine.pdf}
    \par

    \vspace{-4pt}

    {\small Separation sweep\par}
    \vspace{1pt}

    \noindent
    \includegraphics[
        width=0.31\linewidth,
        trim={0mm 2mm 0mm 2mm},
        clip
    ]{separation_scaled_mse.pdf}%
    \hspace{0.005\linewidth}%
    \includegraphics[
        width=0.31\linewidth,
        trim={0mm 2mm 0mm 2mm},
        clip
    ]{separation_scaled_relmse.pdf}%
    \hspace{0.005\linewidth}%
    \includegraphics[
        width=0.31\linewidth,
        trim={0mm 2mm 0mm 2mm},
        clip
    ]{separation_cosine.pdf}
    \par\vspace{2pt}

    \noindent
    \begin{adjustbox}{max width=\linewidth}
    {\footnotesize
    \begin{tikzpicture}[baseline=-0.5ex]
        \draw[toyBOT,line width=0.9pt]
            (0,0)--(0.35,0);
    \end{tikzpicture}%
    \ W-Flow
    \hspace{0.6em}
    \begin{tikzpicture}[baseline=-0.5ex]
        \draw[toyTauTiny,line width=0.9pt,dashed]
            (0,0)--(0.35,0);
    \end{tikzpicture}%
    \ \methodname{}\,($\tau=0.05$)
    \hspace{0.6em}
    \begin{tikzpicture}[baseline=-0.5ex]
        \draw[toyTauSmall,line width=0.9pt,dashdotted]
            (0,0)--(0.35,0);
    \end{tikzpicture}%
    \ \methodname{}\,($\tau=0.1$)
    \hspace{0.6em}
    \begin{tikzpicture}[baseline=-0.5ex]
        \draw[toyTauOne,line width=1pt,densely dotted]
            (0,0)--(0.35,0);
    \end{tikzpicture}%
    \ \methodname{}\,($\tau=1$)
    \hspace{0.6em}
    \begin{tikzpicture}[baseline=-0.5ex]
        \draw[toyTauTen,line width=0.9pt,dashed]
            (0,0)--(0.35,0);
    \end{tikzpicture}%
    \ \methodname{}\,($\tau=10$)
    }
    \end{adjustbox}
    \par

    \caption{\textbf{Full fixed-support velocity-error results.}
    Columns show $B\times\mathrm{MSE}$,
    $B\times\mathrm{RelMSE}$, and cosine similarity.
    The top and bottom rows vary $B$ and $d$, respectively.}
    \label{fig:toy-velocity-fixed-full}
\end{figure*}

For the separation sweep, we fix $B=16$ and use
$d\in\{0.5,1,2,4,8\}$.
We compare UOT with $\tau\in\{0.05,0.1,1,10\}$ against
balanced OT, corresponding to $\tau=\infty$.
\Figref{fig:toy-velocity-fixed-full} presents the full results
for $B\times\mathrm{MSE}$, $B\times\mathrm{RelMSE}$,
and cosine similarity.

\subsection{Continuous Gaussian-Mixture Setting}
\label{app:gaussian-mix-setting}
To move beyond fixed supports, we replace each point above with an
isotropic Gaussian component. Specifically, the source and target
distributions are equal-weight mixtures centered at
\[
(\pm d/2,0)
\qquad\text{and}\qquad
(\pm d/2,h),
\]
respectively, with covariance $\sigma^2 I$. We use $h=1$ and
$\sigma=0.15$. Fresh samples are drawn at every trial, so minibatch
variation affects both the empirical mode proportions and the sampled
support locations.

\begin{figure*}[t]
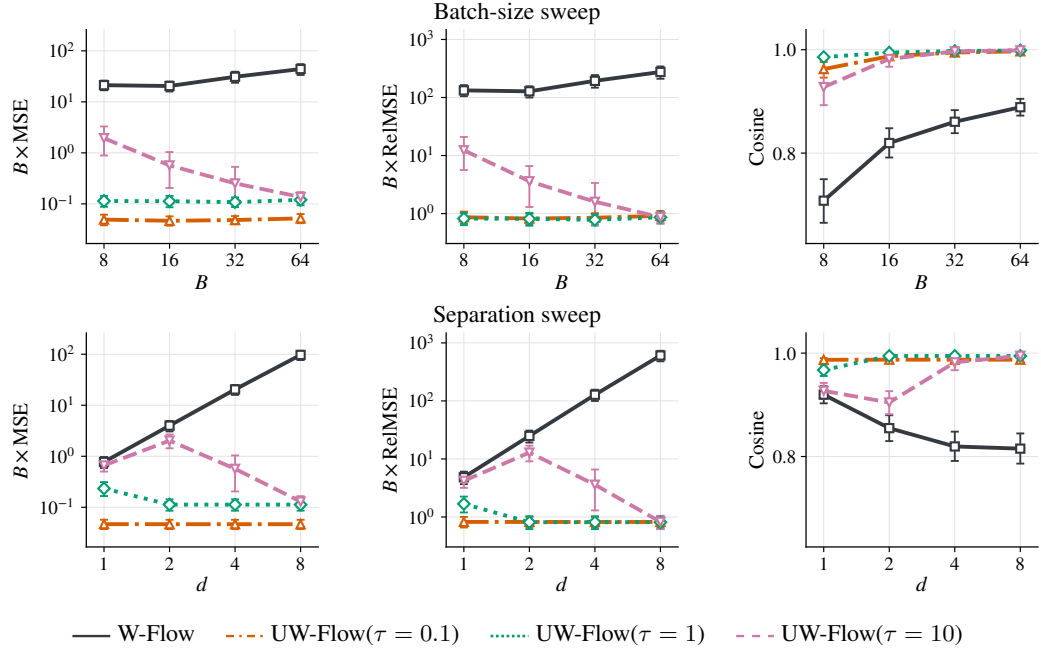

    \centering

    \graphicspath{{figures/toy_velocity_appendix_gaussian/}{toy_velocity_appendix_gaussian/}{./}}

    \definecolor{toyBOT}{HTML}{343A40}
    \definecolor{toyTauSmall}{HTML}{D55E00}
    \definecolor{toyTauOne}{HTML}{009E73}
    \definecolor{toyTauTen}{HTML}{CC79A7}

    {\small Batch-size sweep\par}
    \vspace{1pt}

    \noindent
    \includegraphics[
        width=0.31\linewidth,
        trim={0mm 2mm 0mm 2mm},
        clip
    ]{batch_scaled_mse.pdf}%
    \hspace{0.005\linewidth}%
    \includegraphics[
        width=0.31\linewidth,
        trim={0mm 2mm 0mm 2mm},
        clip
    ]{batch_scaled_relmse.pdf}%
    \hspace{0.005\linewidth}%
    \includegraphics[
        width=0.31\linewidth,
        trim={0mm 2mm 0mm 2mm},
        clip
    ]{batch_cosine.pdf}
    \par

    \vspace{-4pt}

    {\small Separation sweep\par}
    \vspace{1pt}

    \noindent
    \includegraphics[
        width=0.31\linewidth,
        trim={0mm 2mm 0mm 2mm},
        clip
    ]{separation_scaled_mse.pdf}%
    \hspace{0.005\linewidth}%
    \includegraphics[
        width=0.31\linewidth,
        trim={0mm 2mm 0mm 2mm},
        clip
    ]{separation_scaled_relmse.pdf}%
    \hspace{0.005\linewidth}%
    \includegraphics[
        width=0.31\linewidth,
        trim={0mm 2mm 0mm 2mm},
        clip
    ]{separation_cosine.pdf}
    \par\vspace{2pt}

    {\small
    \begin{tikzpicture}[baseline=-0.5ex]
        \draw[toyBOT,line width=0.9pt]
            (0,0)--(0.50,0);
    \end{tikzpicture}%
    \ W-Flow
    \quad
    \begin{tikzpicture}[baseline=-0.5ex]
        \draw[toyTauSmall,line width=0.9pt,dashdotted]
            (0,0)--(0.50,0);
    \end{tikzpicture}%
    \ \methodname{}($\tau=0.1$)
    \quad
    \begin{tikzpicture}[baseline=-0.5ex]
        \draw[toyTauOne,line width=1pt,densely dotted]
            (0,0)--(0.50,0);
    \end{tikzpicture}%
    \ \methodname{}($\tau=1$)
    \quad
    \begin{tikzpicture}[baseline=-0.5ex]
        \draw[toyTauTen,line width=0.9pt,dashed]
            (0,0)--(0.50,0);
    \end{tikzpicture}%
    \ \methodname{}($\tau=10$)
    \par
    }

    \caption{\textbf{Velocity errors under Gaussian-mixture sampling.}
    Columns show $B\times\mathrm{MSE}$,
    $B\times\mathrm{RelMSE}$, and cosine similarity.
    The top and bottom rows vary $B$ and $d$, respectively.}
    \label{fig:toy-velocity-gaussian-full}
\end{figure*}

We use $B\in\{8,16,32,64\}$ at $d=4$ for the batch-size sweep
and $d\in\{1,2,4,8\}$ at $B=16$ for the separation sweep.
We compare UOT with $\tau\in\{0.1,1,10\}$ against balanced OT.
Each configuration is evaluated over 64 independent trials.
\Figref{fig:toy-velocity-gaussian-full} reports the corresponding
$B\times\mathrm{MSE}$, $B\times\mathrm{RelMSE}$,
and cosine similarity.

\section{Experimental Details and Additional Generation Results}
\label{app:experiments}

This appendix first presents the training algorithm and unbalanced Sinkhorn solver (\Secref{app:training_algorithm}). We then provide the setup and full $B\times\tau$ ablations
for toy distribution matching (\Secref{app:experiments_5.1}),
followed by detailed settings and complete results for the
fixed-budget minibatch ablation (\Secref{app:batch_ablation})
and the image-generation hyperparameters
(\Secref{app:hyperparameters}).
We also report ImageNet-100 generation results (\Secref{app:imagenet_100}),
together with ablations of the Sinkhorn iteration count and
entropic coefficient (\Secref{app:additional_ablations}).

\subsection{Training Algorithm}
\label{app:training_algorithm}
\definecolor{uotblue}{RGB}{0,65,190}
\providecommand{\uotchange}[1]{{\color{uotblue}#1}}

\begin{figure*}[t]
\centering
\begingroup
\footnotesize
\algrenewcommand{\algorithmicindent}{0.8em}
\algrenewcommand{\algorithmicrequire}{\textbf{Input:}}

\begin{tabular}{@{}p{0.485\textwidth}@{\hspace{0.03\textwidth}}p{0.485\textwidth}@{}}

\begin{minipage}[t]{\linewidth}
\vspace{0pt}
\hrule height 0.7pt
\vspace{3pt}
\refstepcounter{algorithm}\label{alg:uot_wflow}
\textbf{Algorithm \thealgorithm}\enspace \methodname{} Training
\par\vspace{3pt}
\hrule height 0.35pt
\vspace{4pt}

\begin{algorithmic}[1]
\setlength{\itemsep}{1.0pt}
\Require Generator $f_\theta$; distributions $p_{\rm ref},p$;
sample sizes $N,M$; step size $\eta$;
$\varepsilon,\uotchange{\tau},L$.

\Repeat
    \State Sample $Z=\{z_i\}_{i=1}^{N}\sim p_{\rm ref}$
    \State Sample $Z'=\{z'_i\}_{i=1}^{N}\sim p_{\rm ref}$
    \State Sample $Y=\{y_j\}_{j=1}^{M}\sim p$

    \State $X\gets f_\theta(Z)$
    \State $X'\gets\operatorname{sg}(f_\theta(Z'))$

    \State \uotchange{$
        (T^+,s^+)\gets
        \textsc{USinkhorn}(X,Y;\varepsilon,\tau,L)
    $}
    \State \uotchange{$
        (T^-,s^-)\gets
        \textsc{USinkhorn}(X,X';\varepsilon,\tau,L)
    $}

    \State \uotchange{$
        V_i\gets s_i^+(T_i^+-x_i)-s_i^-(T_i^--x_i)
    $}

    \State $\widetilde{x}_i\gets
        \operatorname{sg}(x_i+\eta V_i)$

    \State Update $\theta$ by minimizing
    \Statex \hspace{\algorithmicindent}
    $
        \mathcal L(\theta)=
        \frac{1}{N}\sum_{i=1}^{N}
        \|f_\theta(z_i)-\widetilde{x}_i\|_2^2
    $
\Until{converged}
\end{algorithmic}
\end{minipage}
&
\begin{minipage}[t]{\linewidth}
\vspace{0pt}
\hrule height 0.7pt
\vspace{3pt}
\refstepcounter{algorithm}\label{alg:uot_sinkhorn}
\textbf{Algorithm \thealgorithm}\enspace Unbalanced Sinkhorn
\par\vspace{3pt}
\hrule height 0.35pt
\vspace{4pt}

\begin{algorithmic}[1]
\setlength{\itemsep}{0.2pt}

\Require Particles $X=\{x_i\}_{i=1}^{N}$,
$Y=\{y_j\}_{j=1}^{M}$; $\varepsilon,\tau,L$.

\State $a_i\gets1/N,\quad b_j\gets1/M$
    \uotnote{Weights}

\State $C_{ij}\gets\tfrac12\|x_i-y_j\|_2^2$
    \uotnote{Cost}

\State $\phi_i\gets0,\quad\psi_j\gets0$
    \uotnote{Initialize}

\State $\alpha\gets\dfrac{\tau}{\tau+\varepsilon}$
    \uotnote{Relaxation}

\For{$\ell=1,\ldots,L$}
    \State $
        \phi_i\gets-\alpha\varepsilon
        \log\sum_j b_j e^{(\psi_j-C_{ij})/\varepsilon}$
        \uotnote{Source}

    \State $
        \psi_j\gets-\alpha\varepsilon
        \log\sum_i a_i e^{(\phi_i-C_{ij})/\varepsilon}$
        \uotnote{Target}
\EndFor

\State $\Pi_{ij}\gets
    a_i b_j e^{(\phi_i+\psi_j-C_{ij})/\varepsilon}$
    \uotnote{Coupling}

\State $ r_i\gets\sum_j\Pi_{ij}$
    \uotnote{Row mass}

\State $s_i\gets\dfrac{r_i}{a_i}$
    \uotnote{Mass ratio}

\State $
    T_i\gets\frac{1}{r_i}\sum_j\Pi_{ij}y_j$
    \uotnote{Barycenter}

\State \Return $(T,s)$
\end{algorithmic}
\end{minipage}
\\[3pt]
\multicolumn{1}{@{}l}{\rule{0.485\textwidth}{0.35pt}}
&
\multicolumn{1}{l@{}}{\rule{0.485\textwidth}{0.35pt}}
\end{tabular}
\endgroup
\end{figure*}

Algorithm~\ref{alg:uot_wflow} summarizes the training procedure of \methodname{}.
At each iteration, we compute cross and self UOT couplings to construct
the mass-weighted velocity field in~\eqref{eq:uot_wflow_velocity}, and update the generator
by regressing toward stop-gradient targets.
Algorithm~\ref{alg:uot_sinkhorn} details the unbalanced Sinkhorn
solver, which returns the barycentric projections and transported-mass
ratios used to construct this field.

\subsection{Generation on Synthetic Distributions}
\label{app:experiments_5.1}
We provide experimental details and full $B\times\tau$ ablations
for the synthetic generation experiments in \Secref{sec:toy_results}.
We consider three two-dimensional target distributions:
checkerboard, two moons, and Gaussian ring.

\textbf{Training configuration.}
W-Flow and \methodname{} use the same Fourier-feature generator with an effective optimizer batch size of $B_{\mathrm{opt}}=512$. We vary the transport batch size $B\in\{8,32,128,512\}$ and evaluate \methodname{} with $\tau\in\{0.01,0.1,1,10\}$. All runs use 30,000 optimizer updates, 30 Sinkhorn iterations, $\varepsilon=0.01$, and random seed 0. The reported $\varepsilon$ and $\tau$ denote coefficients in the normalized training loss.

\begin{figure*}[p]
    \centering
    \includegraphics[width=\textwidth]{%
        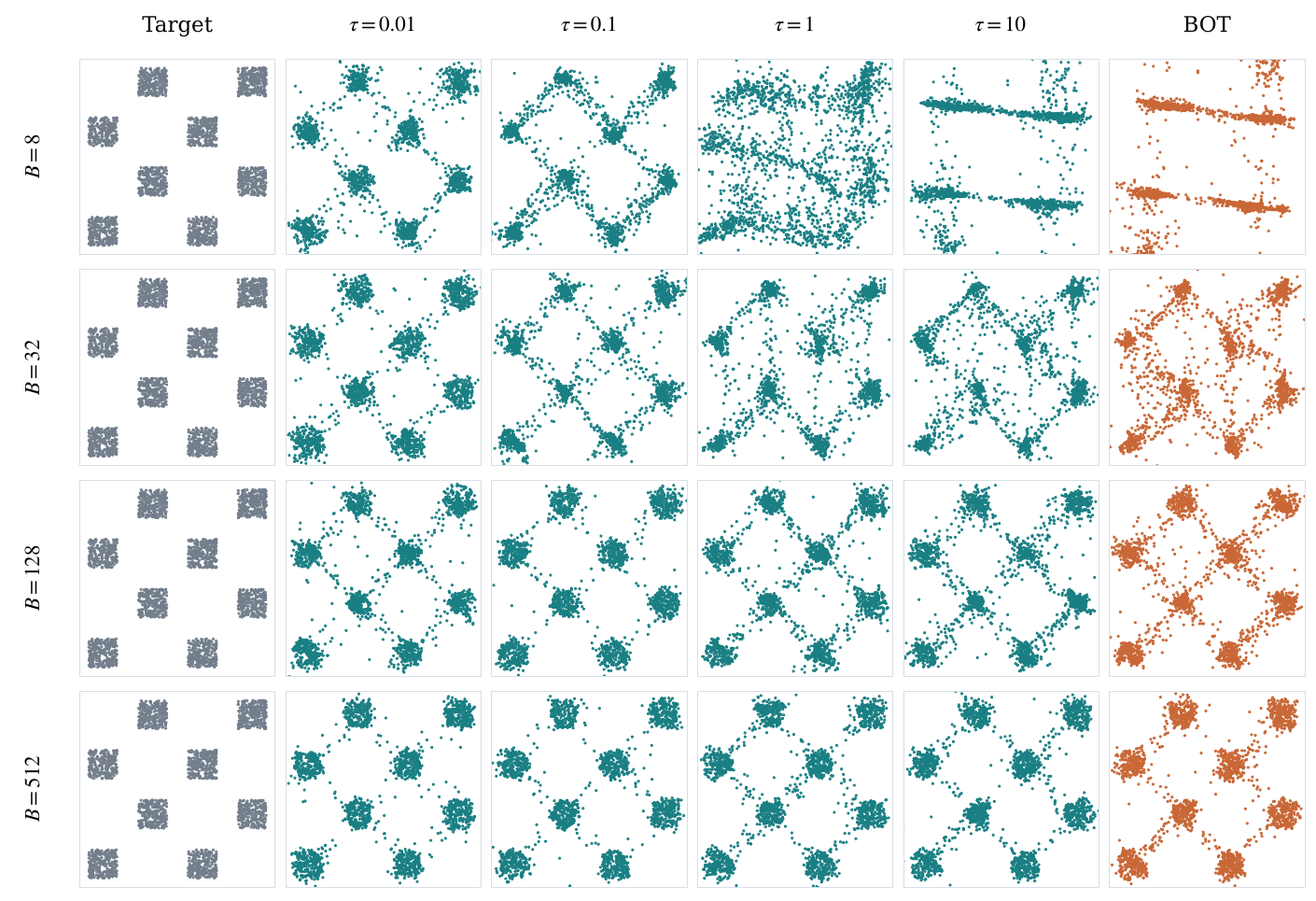%
    }
    \caption{
        Qualitative $B\times\tau$ ablation on checkerboard.
        Rows vary $B$; columns show the target,
        \methodname{} with increasing $\tau$, and W-Flow.
    }
    \label{fig:toy_checkerboard_B_tau}
\end{figure*}

\begin{figure*}[p]
    \centering
    \includegraphics[width=\textwidth]{%
        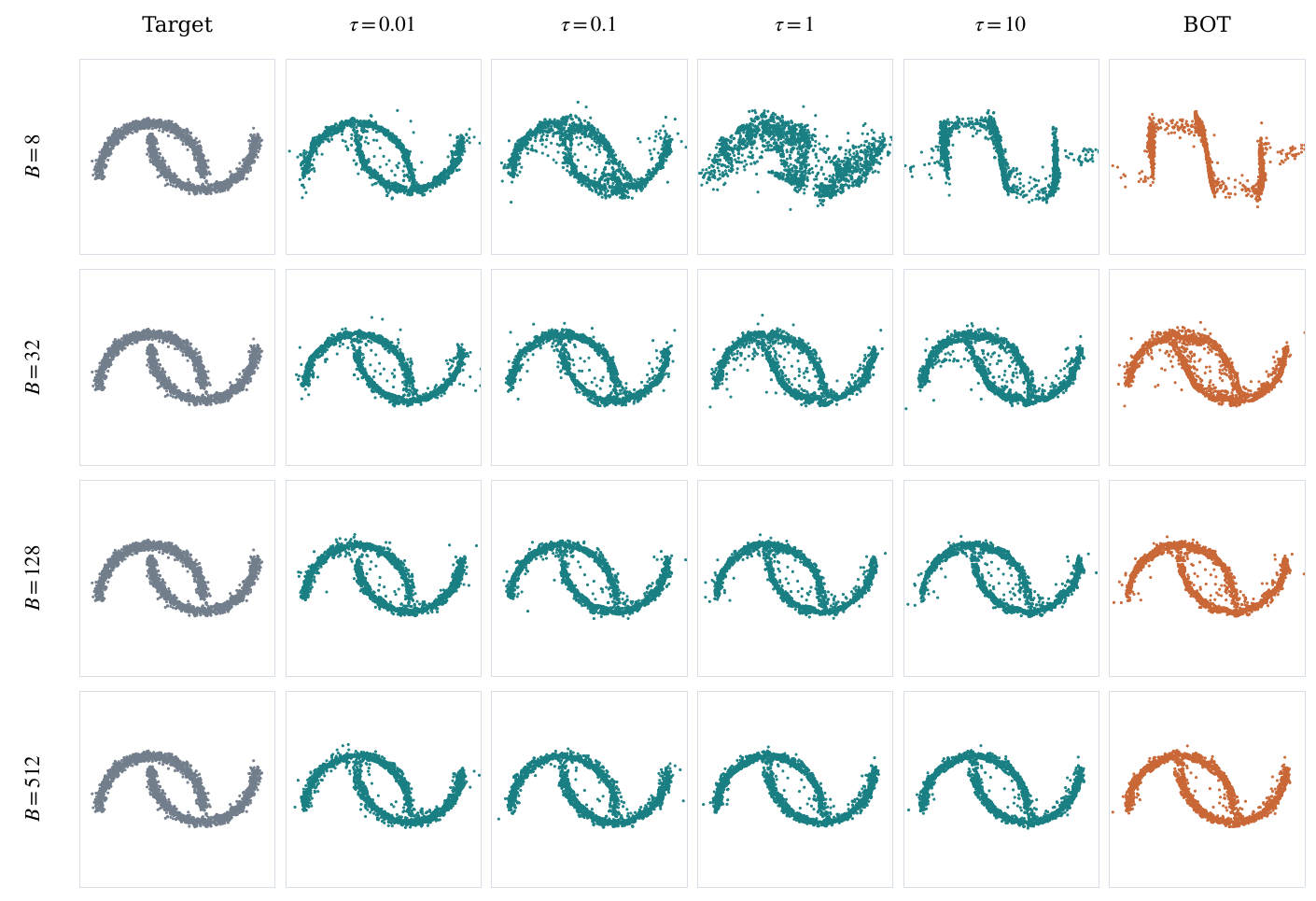%
    }
    \caption{
        Qualitative $B\times\tau$ ablation on two moons.
        Rows vary $B$; columns show the target,
        \methodname{} with increasing $\tau$, and W-Flow.
    }
    \label{fig:toy_two_moons_B_tau}
\end{figure*}

\begin{figure*}[p]
    \centering
    \includegraphics[width=\textwidth]{%
        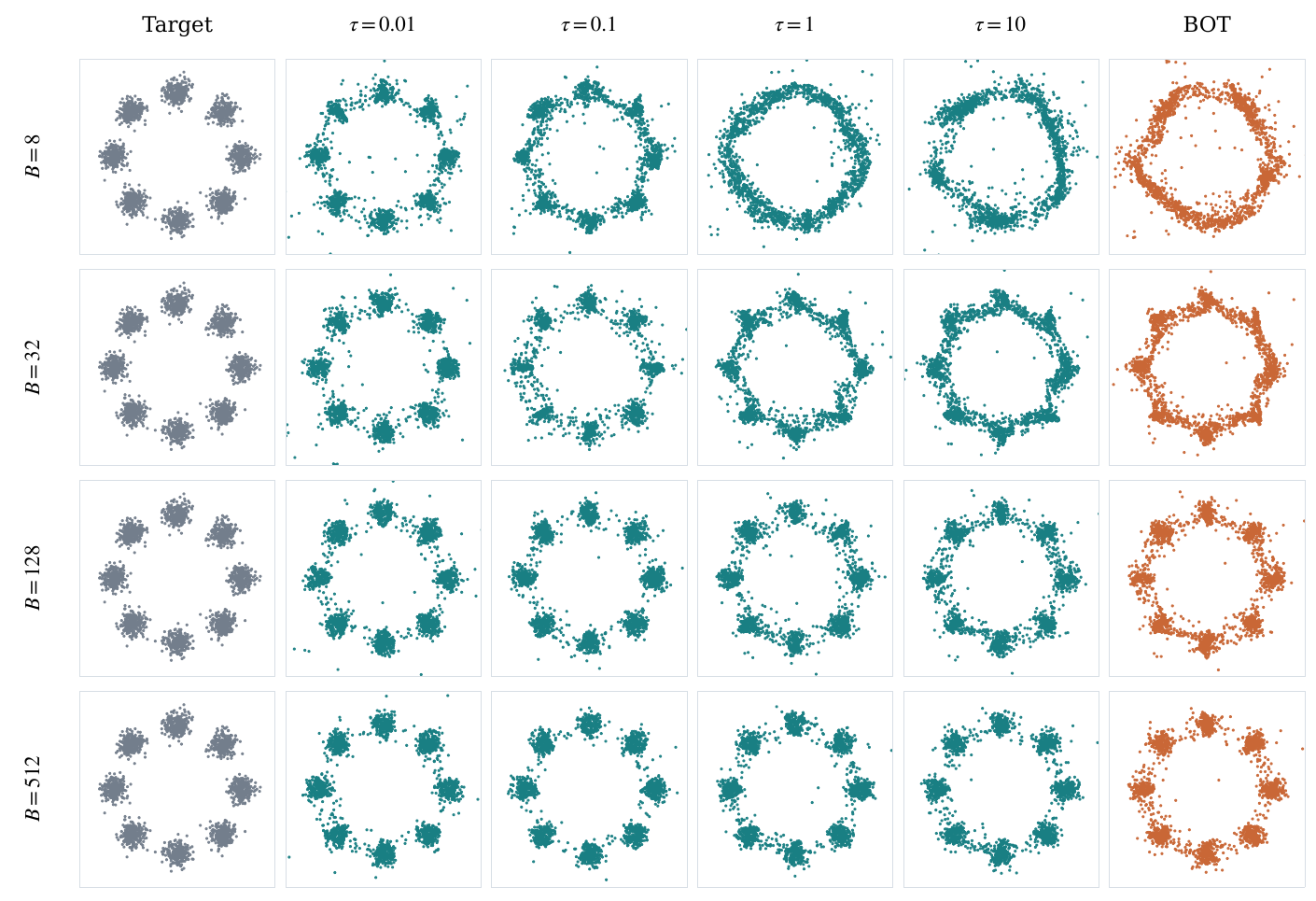%
    }
    \caption{
        Qualitative $B\times\tau$ ablation on Gaussian ring.
        Rows vary $B$; columns show the target,
        \methodname{} with increasing $\tau$, and W-Flow.
    }
    \label{fig:toy_gaussian_ring_B_tau}
\end{figure*}

\textbf{Full qualitative ablations.} \Figref{fig:toy_checkerboard_B_tau},
\Figref{fig:toy_two_moons_B_tau}, and
\Figref{fig:toy_gaussian_ring_B_tau}
show that \methodname{} with $\tau\in\{0.01,0.1\}$
generally captures the target structures more clearly than
W-Flow at smaller $B$.
As $B$ increases, both methods improve and the qualitative
gap tends to narrow.
Larger $\tau$ generally produces results closer to W-Flow.

\subsection{Fixed-Budget Minibatch Ablation}
\label{app:batch_ablation}

Each transport group contains $B$ source samples, $B$ target
samples, and $B$ independently drawn self-reference samples.
Cross and self transport couplings are computed separately
within each group.

We vary $B\in\{4,8,16,32,64\}$ while keeping
\[
B_{\mathrm{opt}}=N_cB
\]
fixed at $64$ for MNIST and $128$ for CIFAR-10.
Accordingly, we set $N_c=B_{\mathrm{opt}}/B$.
The corresponding values of $N_c$ are
$\{16,8,4,2,1\}$ for MNIST and
$\{32,16,8,4,2\}$ for CIFAR-10.
Regression losses from all groups are aggregated
for each optimizer update.

For each dataset and support size, W-Flow and UW-Flow
use the same model, feature representation, optimizer,
training schedule, and entropic regularization parameter.
They differ in the transport objective used to construct
the velocity.
We compute the evaluation metrics using 50,000 generated
samples for CIFAR-10 and 10,000 for MNIST.
Additional hyperparameters and evaluation settings are
provided in Appendix~\ref{app:hyperparameters}.
The complete results corresponding to
\Tabref{tab:mnist_support_ablation} and
\Tabref{tab:cifar10_support_ablation} are reported in
\Tabref{tab:app_mnist_support_full} and
\Tabref{tab:app_cifar10_support_full}, respectively.

\begin{table}[t]
\centering
\caption{Full results of the fixed-budget minibatch discrepancy ablation on unconditional MNIST, with fixed $B_{\mathrm{opt}}=N_cN_{\mathrm{gen}}=64$.}
\label{tab:app_mnist_support_full}
\small
\setlength{\tabcolsep}{5pt}
\begin{tabular}{@{}ccclccc@{}}
\toprule
$N_c$ & $N_{\mathrm{gen}}$ & $N_{\mathrm{pos}}$ & Method & FID $\downarrow$ & Prec. $\uparrow$ & Rec. $\uparrow$ \\
\midrule
\multirow{4}{*}{16} & \multirow{4}{*}{4} & \multirow{4}{*}{4} & W-Flow & 19.70 & 0.536 & 0.059 \\
& & & \methodname{} ($\tau=10$) & 22.54 & 0.562 & 0.081 \\
& & & \methodname{} ($\tau=1$) & 17.98 & 0.464 & 0.116 \\
& & & \methodname{} ($\tau=0.1$) & 18.66 & 0.419 & 0.253 \\
\midrule
\multirow{4}{*}{8} & \multirow{4}{*}{8} & \multirow{4}{*}{8} & W-Flow & 13.69 & 0.461 & 0.239 \\
& & & \methodname{} ($\tau=10$) & 13.50 & 0.545 & 0.110 \\
& & & \methodname{} ($\tau=1$) & 12.37 & 0.477 & 0.281 \\
& & & \methodname{} ($\tau=0.1$) & 13.56 & 0.492 & 0.351 \\
\midrule
\multirow{4}{*}{4} & \multirow{4}{*}{16} & \multirow{4}{*}{16} & W-Flow & 9.86 & 0.487 & 0.430 \\
& & & \methodname{} ($\tau=10$) & 10.14 & 0.513 & 0.355 \\
& & & \methodname{} ($\tau=1$) & 9.97 & 0.517 & 0.218 \\
& & & \methodname{} ($\tau=0.1$) & 9.22 & 0.470 & 0.507 \\
\midrule
\multirow{4}{*}{2} & \multirow{4}{*}{32} & \multirow{4}{*}{32} & W-Flow & 8.72 & 0.576 & 0.317 \\
& & & \methodname{} ($\tau=10$) & 7.33 & 0.596 & 0.343 \\
& & & \methodname{} ($\tau=1$) & 7.85 & 0.517 & 0.435 \\
& & & \methodname{} ($\tau=0.1$) & 7.78 & 0.528 & 0.530 \\
\midrule
\multirow{4}{*}{1} & \multirow{4}{*}{64} & \multirow{4}{*}{64} & W-Flow & 7.03 & 0.572 & 0.532 \\
& & & \methodname{} ($\tau=10$) & 6.11 & 0.589 & 0.578 \\
& & & \methodname{} ($\tau=1$) & 5.67 & 0.585 & 0.568 \\
& & & \methodname{} ($\tau=0.1$) & 6.18 & 0.559 & 0.603 \\
\bottomrule
\end{tabular}
\end{table}

\begin{table}[t]
\centering
\caption{Full results of the fixed-budget minibatch discrepancy ablation on unconditional CIFAR-10, with fixed $B_{\mathrm{opt}}=N_cN_{\mathrm{gen}}=128$.}
\label{tab:app_cifar10_support_full}
\small
\setlength{\tabcolsep}{4pt}
\begin{tabular}{@{}ccclcccc@{}}
\toprule
$N_c$ & $N_{\mathrm{gen}}$ & $N_{\mathrm{pos}}$ & Method & FID $\downarrow$ & IS $\uparrow$ & Prec. $\uparrow$ & Rec. $\uparrow$ \\
\midrule
\multirow{4}{*}{32} & \multirow{4}{*}{4} & \multirow{4}{*}{4} & W-Flow & 30.58 & 9.75 & 0.55 & 0.01 \\
& & & \methodname{} ($\tau=10$) & 22.03 & 9.58 & 0.57 & 0.02 \\
& & & \methodname{} ($\tau=1$) & 19.41 & 10.10 & 0.56 & 0.02 \\
& & & \methodname{} ($\tau=0.1$) & 15.56 & 10.60 & 0.52 & 0.07 \\
\midrule
\multirow{4}{*}{16} & \multirow{4}{*}{8} & \multirow{4}{*}{8} & W-Flow & 24.39 & 8.99 & 0.58 & 0.01 \\
& & & \methodname{} ($\tau=10$) & 14.64 & 10.39 & 0.59 & 0.04 \\
& & & \methodname{} ($\tau=1$) & 13.71 & 10.82 & 0.56 & 0.05 \\
& & & \methodname{} ($\tau=0.1$) & 9.70 & 10.53 & 0.56 & 0.17 \\
\midrule
\multirow{4}{*}{8} & \multirow{4}{*}{16} & \multirow{4}{*}{16} & W-Flow & 12.49 & 10.09 & 0.59 & 0.07 \\
& & & \methodname{} ($\tau=10$) & 10.39 & 10.16 & 0.59 & 0.13 \\
& & & \methodname{} ($\tau=1$) & 10.03 & 9.87 & 0.61 & 0.15 \\
& & & \methodname{} ($\tau=0.1$) & 13.97 & 10.09 & 0.52 & 0.24 \\
\midrule
\multirow{4}{*}{4} & \multirow{4}{*}{32} & \multirow{4}{*}{32} & W-Flow & 9.57 & 10.32 & 0.63 & 0.08 \\
& & & \methodname{} ($\tau=10$) & 8.68 & 10.19 & 0.61 & 0.23 \\
& & & \methodname{} ($\tau=1$) & 8.43 & 10.20 & 0.60 & 0.16 \\
& & & \methodname{} ($\tau=0.1$) & 5.01 & 10.43 & 0.61 & 0.39 \\
\midrule
\multirow{4}{*}{2} & \multirow{4}{*}{64} & \multirow{4}{*}{64} & W-Flow & 8.10 & 10.37 & 0.60 & 0.27 \\
& & & \methodname{} ($\tau=10$) & 6.67 & 10.30 & 0.60 & 0.31 \\
& & & \methodname{} ($\tau=1$) & 5.11 & 10.23 & 0.62 & 0.39 \\
& & & \methodname{} ($\tau=0.1$) & 4.51 & 10.37 & 0.60 & 0.45 \\
\bottomrule
\end{tabular}
\end{table}

\subsection{Image-Generation Hyperparameters}
\label{app:hyperparameters}

\Tabref{tab:hyperparameters} reports the model, transport, and optimization settings for the MNIST~\citep{lecun:1998mnist} and CIFAR-10~\citep{krizhevsky:2009cifar10} experiments in \Secref{sec:batch_ablation} and the ImageNet-100~\citep{deng:2009imagenet} experiments in \Secref{sec:imagenet100}. The ImageNet-100 experiments are based on the ImageNet ablation setup of W-Flow~\citep{han:2026wflow}. For the fixed-budget ablation, we vary $N_c$ and the samples per transport group as detailed in Appendix~\ref{app:batch_ablation}.

For \Secref{sec:batch_ablation}, we vary $N_c$ and the number of samples per transport group while fixing the per-update budget $B_{\mathrm{opt}}$, as detailed in Appendix~\ref{app:batch_ablation}. The other dataset-specific hyperparameters follow \Tabref{tab:hyperparameters} unless stated otherwise. For ImageNet-100, we follow the ImageNet ablation setup of W-Flow~\citep{han:2026wflow}, changing only the number of classes to 100.

\begin{table}[t]
    \centering
    \caption{Hyperparameters for image-generation experiments. $N_{\mathrm{gen}}$ and $N_{\mathrm{pos}}$ are the numbers of generated and target samples per transport group; the total generated batch size is $N_cN_{\mathrm{gen}}$. The listed $\varepsilon$ and $\tau$ values are before scaling by the feature dimension $D$.}
    \label{tab:hyperparameters}
    \small
    \setlength{\tabcolsep}{4pt}
    \renewcommand{\arraystretch}{1.2}
    \begin{tabularx}{\linewidth}{
        @{}l|
        >{\centering\arraybackslash}X|
        >{\centering\arraybackslash}X|
        >{\centering\arraybackslash}X@{}
    }
        \toprule
        Hyperparameter & MNIST & CIFAR-10 & ImageNet-100 \\
        \midrule
        \rowcolor{gray!15}
        \multicolumn{4}{@{}l}{\textbf{Model and representation}} \\
        \addlinespace[2pt]
        Generator
            & DiT-based, patch 2
            & DiT-based, patch 2
            & DiT-B/2 \\
        Training shape / space
            & $28\!\times\!28\!\times\!1$
            & $32\!\times\!32\!\times\!3$
            & $32\!\times\!32\!\times\!4$ \\
        Feature encoder (frozen)
            & ConvNeXt V2--Tiny
            & ConvNeXt V2--Tiny
            & Latent-MAE ResNet\newline (width 256) \\
        Feature normalization
            & Distance-based scaling
            & Distance-based scaling
            & Distance-based scaling \\
        Velocity normalization
            & RMS normalization
            & RMS normalization
            & RMS normalization \\
        \midrule
        \rowcolor{gray!15}
        \multicolumn{4}{@{}l}{\textbf{Transport}} \\
        \addlinespace[2pt]
        Per-update budget $B_{\mathrm{opt}} = N_c N_q$
            & 64 & 128 & 4,096 \\
        Samples per group $N_{\mathrm{gen}} = N_{\mathrm{pos}}$\,$^{\dagger}$
            & $\{4,8,16,32,64\}$
            & $\{4,8,16,32,64\}$
            & 64 \\
        Transport groups $N_c$
            & $B_{\mathrm{opt}}/N_{\mathrm{gen}}$
            & $B_{\mathrm{opt}}/N_{\mathrm{gen}}$
            & 64 \\
        Entropic coefficient $\varepsilon$
            & 0.05 & 0.05 & 0.05 \\
        Marginal penalty coefficient $\tau$
            & $\{0.1,1,10\}$
            & $\{0.1,1,10\}$
            & $\{0.1,1,10\}$ \\
        Sinkhorn iterations $L$
            & 30 & 30 & 10 \\
        Drift step size $\eta$
            & 1.0 & 1.0 & 1.0 \\
        \midrule
        \rowcolor{gray!15}
        \multicolumn{4}{@{}l}{\textbf{Optimization}} \\
        \addlinespace[2pt]
        Optimizer
            & AdamW & AdamW & AdamW \\
        $(\beta_1,\beta_2)$
            & $(0.9,0.95)$
            & $(0.9,0.95)$
            & $(0.9,0.95)$ \\
        Learning rate
            & $2\!\times\!10^{-4}$
            & $2\!\times\!10^{-4}$
            & $2\!\times\!10^{-4}$ \\
        Weight decay
            & 0.01 & 0.01 & 0.01 \\
        LR schedule (after warmup)
            & Cosine & Cosine & Constant \\
        EMA decay
            & 0.999 & 0.999 & 0.999 \\
        Training budget
            & 100 epochs & 500 epochs & 100 epochs \\
        \bottomrule
    \end{tabularx}
    {\footnotesize $^{\dagger}$For MNIST and CIFAR-10, $N_{\mathrm{gen}}$ and $N_c$ are varied jointly
with $B_{\mathrm{opt}}$ fixed. \par}
\end{table}

\subsection{ImageNet-100 Generation}
\label{app:imagenet_100}

\paragraph{Dataset and training setup.} We use the 100 ImageNet classes specified by the ImageNet-100 class list released with Contrastive Multiview Coding~\citep{tian:2020cmc}. We follow the ImageNet ablation setup of W-Flow~\citep{han:2026wflow}, adapting it to the 100-class subset. We train a DiT-B/2 generator in the SD-VAE~\citep{rombach:2022ldm} latent space and compute transport costs using features from a frozen ResNet~\citep{he:2016resnet} encoder pretrained with latent-MAE. Both methods use the same architecture, encoder weights, and optimization protocol within each setting. We use $N_c=64$ transport groups with $N_{\mathrm{gen}}=N_{\mathrm{pos}}=64$ samples per group, giving $B_{\mathrm{opt}}=N_cN_{\mathrm{gen}}=4096$. Additional hyperparameters are listed in \Tabref{tab:hyperparameters}.

\paragraph{Parameter selection and evaluation.} We evaluate \methodname{} with $\tau\in\{0.1,1,10\}$ in both settings. For class-conditional evaluation, we follow W-Flow's CFG selection protocol, testing guidance scales in $\{1.0,1.5,2.0,2.5,3.0,3.5,4.0\}$ and reporting the metrics at the scale with the lowest FID for each method. \Tabref{tab:app_imagenet100_full} reports the complete results across $\tau$.

\begin{table*}[t]
\centering
\caption{Full ImageNet-100 generation results following the ImageNet ablation setup and CFG selection protocol of W-Flow~\citep{han:2026wflow}. Both settings use $N_c=64$, $N_{\mathrm{gen}}=N_{\mathrm{pos}}=64$, and $B_{\mathrm{opt}}=4096$. For class-conditional generation, we select the CFG scale with the lowest FID for each method from $\{1.0,1.5,2.0,2.5,3.0,3.5,4.0\}$.}
\label{tab:app_imagenet100_full}
\small
\setlength{\tabcolsep}{4pt}
\begin{tabular}{@{}lcccclcccc@{}}
\toprule
Setting & $N_c$ & $N_{\mathrm{gen}}$ & $N_{\mathrm{pos}}$ & $B_{\mathrm{opt}}$ & Method & FID $\downarrow$ & IS $\uparrow$ & Prec. $\uparrow$ & Rec. $\uparrow$ \\
\midrule
\multirow{4}{*}{Uncond} & \multirow{4}{*}{64} & \multirow{4}{*}{64} & \multirow{4}{*}{64} & \multirow{4}{*}{4096} & W-Flow & 55.25 & 23.89 & 0.42 & 0.14 \\
& & & & & \methodname{} ($\tau=10$) & 55.00 & 25.01 & 0.40 & 0.16 \\
& & & & & \methodname{} ($\tau=1$) & 51.08 & 25.47 & 0.41 & 0.19 \\
& & & & & \methodname{} ($\tau=0.1$) & 46.17 & 27.71 & 0.40 & 0.29 \\
\midrule
\multirow{4}{*}{Cond} & \multirow{4}{*}{64} & \multirow{4}{*}{64} & \multirow{4}{*}{64} & \multirow{4}{*}{4096} & W-Flow & 9.43 & 60.31 & 0.74 & 0.26 \\
& & & & & \methodname{} ($\tau=10$) & 9.47 & 61.08 & 0.75 & 0.25 \\
& & & & & \methodname{} ($\tau=1$) & 9.11 & 61.34 & 0.75 & 0.26 \\
& & & & & \methodname{} ($\tau=0.1$) & 9.59 & 58.35 & 0.78 & 0.25 \\
\bottomrule
\end{tabular}
\end{table*}

\subsection{Sinkhorn Parameter Ablations}
\label{app:additional_ablations}

\begin{table*}[t]
    \centering
    \caption{
        Sinkhorn parameter ablations on CIFAR-10.
        We fix $N_c=2$, $N_{\mathrm{gen}}=N_{\mathrm{pos}}=64$,
        $B_{\mathrm{opt}}=128$, and $\tau=0.1$.
        The default configuration uses $L=30$ and $\varepsilon=0.05$;
        each table varies one parameter while fixing the other.
    }
    \label{tab:sinkhorn_ablation}

    \small
    \setlength{\tabcolsep}{4pt}

    \begin{minipage}[t]{0.49\textwidth}
        \centering
        \begin{tabular*}{\linewidth}{@{\extracolsep{\fill}}ccccc@{}}
            \toprule
            \multicolumn{5}{c}{\textbf{Sinkhorn iterations $L$}} \\
            \midrule
            $L$ & FID $\downarrow$ & IS $\uparrow$
            & Precision $\uparrow$ & Recall $\uparrow$ \\
            \midrule
            $5$  & 5.04 & 10.39 & 0.57 & 0.41 \\
            $10$ & 4.73 & \textbf{10.54} & 0.58 & \textbf{0.45} \\
            $20$ & 5.16 & 10.26 & 0.59 & 0.40 \\
            $30$ & \textbf{4.51} & 10.37 & \textbf{0.60} & \textbf{0.45} \\
            $40$ & 5.18 & 10.29 & 0.57 & \textbf{0.45} \\
            \bottomrule
        \end{tabular*}
    \end{minipage}
    \hfill
    \begin{minipage}[t]{0.49\textwidth}
        \centering
        \begin{tabular*}{\linewidth}{@{\extracolsep{\fill}}ccccc@{}}
            \toprule
            \multicolumn{5}{c}{\textbf{Entropic regularization $\varepsilon$}} \\
            \midrule
            $\varepsilon$ & FID $\downarrow$ & IS $\uparrow$
            & Precision $\uparrow$ & Recall $\uparrow$ \\
            \midrule
            $0.01$  & 3.77 & 10.21 & 0.60 & \textbf{0.52} \\
            $0.025$ & \textbf{3.75} & 10.39 & \textbf{0.61} & \textbf{0.52} \\
            $0.05$  & 4.51 & 10.37 & 0.60 & 0.45 \\
            $0.075$ & 5.47 & \textbf{10.60} & 0.57 & 0.39 \\
            $0.1$   & 8.35 & 10.54 & 0.54 & 0.23 \\
            \bottomrule
        \end{tabular*}
    \end{minipage}
\end{table*}

Following W-Flow~\citep{han:2026wflow}, we examine the effects of Sinkhorn iterations $L$ and the entropic coefficient $\varepsilon$ on UW-Flow. We use the CIFAR-10 configuration with $N_c=2$, $N_{\mathrm{gen}}=N_{\mathrm{pos}}=64$, and $\tau=0.1$, varying one parameter at a time. W-Flow selects $\varepsilon=0.05$ based on its ablation study. We retain this value for the results reported in the main text to compare W-Flow and UW-Flow under the same entropic regularization; the default iteration count is $L=30$.

As shown in \Tabref{tab:sinkhorn_ablation}, FID ranges from $4.51$ to $5.18$ across $L\in\{5,10,20,30,40\}$, with no monotonic improvement as $L$ increases. The choice of $\varepsilon$ has a larger effect in this setting: at $L=30$, reducing $\varepsilon$ from $0.05$ to $0.025$ improves FID from $4.51$ to $3.75$ and recall from $0.45$ to $0.52$. The main-text results use $\varepsilon=0.05$ for the matched comparison; the lower FID at $\varepsilon=0.025$ is reported here as an additional ablation result.

\clearpage
\clearpage

\clearpage

\clearpage

\end{document}